\documentclass{article}
\usepackage[margin=1in]{geometry}
\usepackage{graphicx} 
\usepackage{url}
\usepackage{hyperref}
\usepackage{amsmath}
\usepackage{amsfonts}
\usepackage{amssymb}
\usepackage{amsthm}
\usepackage{circuitikz}
\usepackage{svg}
\usepackage{cleveref}

\theoremstyle{definition}
\newtheorem{definition}{Definition}
\newtheorem{theorem}{Theorem}
\newtheorem{lemma}{Lemma}
\newtheorem{corollary}{Corollary}
\newtheorem{proposition}{Proposition}

\newtheorem{remark}{Remark}
\newtheorem{hypothesis}{Hypothesis}
\crefname{hypothesis}{Hypothesis}{Hypotheses}

\newcommand{\supp}{\operatorname{supp}}
\newcommand{\E}{\mathbb{E}}
\newcommand{\R}{\mathbb{R}}
\newcommand{\Prob}{\mathbb{P}}

\title{Ensembles of Random Partition-Based Models under Covariate Shift\footnote{Preprint. Under review.}}
\author{Mary Versa Clemens-Sewall and Eliza O'Reilly}

\begin{document}

\maketitle

\begin{abstract}
    Distribution shifts, including covariate shifts, are ubiquitous in real-world applications of machine learning. Empirical evidence suggests that, across a variety of models with and without downstream adaptation, ensembles tend to be more robust to covariate shift than single models. We extend the theory of covariate shift severity from single models to ensembles, providing theoretical justification for the empirical trends. We introduce new discrepancy measures of covariate shift severity to control the risk of partition-based kernel estimators. We then show that with sufficient ensemble size, these models can achieve minimax-optimal risk convergence rates under covariate shift. Our results suggest that modelers can expect to need larger ensemble sizes for more severe covariate shifts.
\end{abstract}

\section{Introduction}

Distribution shift occurs when some or all of the data used to train a model is sampled from a \emph{shifted} distribution, meaning one that differs from the distribution of a randomly sampled test point, and it is common in a wide range of machine learning applications \cite{Koh_Sagawa_Marklund_Xie_Zhang_Balsubramani_Hu_Yasunaga_Phillips_Gao_2021}. General distribution shift narrows to covariate shift if the nature of the shift is restricted to the features, a.k.a. covariates. Specifically, covariate shift occurs when the marginal distributions of the feature vectors differ yet the conditional distributions of the labels given the features are identical \cite{quinonero-candela_dataset_2008}. 

Recent work has characterized covariate shifts severe enough that they affect convergence rates of nonparametric estimators, yet not so severe that they preclude learning altogether \cite{kpotufe_marginal_2021,Galbraith_Kpotufe_2024,pathak_new_2022}. These studies have defined asymmetric discrepancy measures for source-target pairs of distributions and proven that these discrepancy measures control minimax rates of convergence. Coarser discrepancy measures, especially the transfer exponent \cite{kpotufe_marginal_2021}, preceded finer ones, like the aggregate transfer exponent \cite{Galbraith_Kpotufe_2024} and integrated transfer exponent \cite{pathak_new_2022,Galbraith_Kpotufe_2024}. 

To establish that these covariate shift discrepancy measures control minimax rates, these works have demonstrated that certain single-learner (i.e., not ensembled) local averaging estimators achieve the reported rates, including nearest-neighbor, dyadic tree, and top-hat (a.k.a. na\"ive \cite{gyorfi_distribution-free_2002}) kernel estimators \cite{kpotufe_marginal_2021,Galbraith_Kpotufe_2024,pathak_new_2022}. In this paper, we extend this theory to random partition-based kernel estimators. 
Our interest in these models stems, in part, from the empirical success of random forest models \cite{Breiman_2001} on many tabular data machine learning problems. Both random forests and random-partition based kernel estimators are ensembles of local averaging estimators where each local averaging estimator is a piecewise constant function over a randomly generated partition. By studying random partition-based kernel estimators, we aim to discover when covariate shift theory developed on single learners applies to models suited to ensembling, and when it does not.

Inspired by the integrated transfer exponent \cite{pathak_new_2022,Galbraith_Kpotufe_2024}, we define a covariate shift discrepancy measure (\cref{df:integrated_P-transfer_discrepancy}) that likewise emerges in the analysis of the estimator's variance. Our random partition-based discrepancy measure exhibits some advantageous theoretical properties (see \cref{pr:integrated_equals_aggregate_P-transfer_discrepancy}). Of practical concern, it cannot always be controlled, even when the integrated transfer exponent can be (see \cref{pr:insufficient_ensemble_size}). Our analysis of a class of doubling measures demonstrates that to control a partition-based estimator's variance under severe covariate shift, ensembling may be necessary, with larger ensembles needed to achieve minimax-optimal rates on more severe covariate shifts.

Our discrepancy measure is specified to the partitioning scheme used to initialize the single learners in the ensemble. We must choose a partitioning scheme to derive risk convergence rates for the random partition-based kernel estimator under covariate shift. We choose to study the STIT tessellation as a partitioning scheme because of its theoretical advantages.

The STIT tree and its ensembles, especially the STIT random forest, have been proven to achieve minimax-optimal convergence rates on Lipshitz regression without covariate shift \cite{oreilly_minimax_2024}. Indeed, the STIT random forest is the only random forest model to date to have been proven minimax-optimal in arbitrary dimension. This was possible because the stochastic geometry of the STIT tessellation can be described with precision \cite{Mourtada_Gaiffas_Scornet_2020,oreilly_minimax_2024}, thanks in part to the memorylessness of the exponential distribution and the Poisson process built upon it. However, this same memorylessness gives the stochastic geometry of the tessellation a heavy-tailed (i.e., exponential-tailed) phenomenon that renders the typical cell in a partition quite small. In a recent work, the authors remarked that the small-cell phenomenon complicates the pursuit of statistical inference on Mondrian trees \cite{Cattaneo_Klusowski_Underwood_2025}, which are a special case of STIT trees. So, too, in our work, the cells of a single Mondrian tree seem to be too small for the single learner to be robust to severe covariate shift (see \cref{pr:insufficient_ensemble_size}). As with statistical inference \cite{Cattaneo_Klusowski_Underwood_2025}, we overcome this challenge through ensembling Mondrian trees to sufficiently widen the pool of independent training points that contribute to the local average.

We prove that, with a sufficient ensemble size, the Mondrian kernel \cite{3020948.3020953} estimator can achieve minimax-optimal rates on a class of severe covariate shifts (see \cref{pr:sufficient_ensemble_size}), just like the top-hat kernel estimator can \cite{pathak_new_2022}. In doing so, we extend covariate shift theory from simpler, single learners to more complex, ensemble estimators.\footnote{Our results are motivated by the study of small ensembles. Better performance guarantees for large ensembles may be obtainable by analyzing the large-ensemble limiting kernel \cite{Lopes_Wu_Lee_2020}, which is the Laplace kernel \cite{3020948.3020953}, but the robustness of Laplace kernel-based models to covariate shift is, to the best of our knowledge, unknown. Indeed, certain partition-based kernel estimators (there termed ``kernel based on random forest'' models) have been studied by comparing to the infinite ensemble-size limit \cite{Scornet_2016}, albeit without considering distribution shift. Our results offer insight into the role of ensemble size, while focusing on the case of small ensemble size.}

\subsection{Empirical Evidence for the Robustness of Ensembles to Covariate Shift}

Empirical studies in the literature report that ensembles outperform single learners under covariate shift. Among popular classification algorithms, random forests consistently experienced less performance degradation under covariate shift than single learners on a wide range of two-dimensional datasets \cite{Bayram_Ahmed_2023}. While individual trees may learn associations that fail to generalize under covariate shift, ensembling them appears to dilute the effect of any one such bias \cite{Bayram_Ahmed_2023,Clark_Yatskar_Zettlemoyer_2019}. 

Many works incorporate ensembling with other interventions to improve the empirical robustness of various models to covariate shift \cite{Clark_Yatskar_Zettlemoyer_2019,Raza_Rathee_Zhou_Cecotti_Prasad_2019,zhang2025novel,3295222.3295387}. The results of these studies show that ensembling benefits performance under covariate shift both with and without additional interventions. For instance, image corruption can induce a covariate shift, and Table 1 in \cite{3540261.3540332} reports that ensembles of Bayesian models outperform single models at multiple levels of corruption. Ensembling helped both baseline neural networks and ones that had been adapted to target data after training. Similar trends are reported on a dataset of financial loss given default, where covariate shift is observed between historical and more recent data. Tables 4 and 5 in \cite{zhang2025novel} demonstrate that for all studied models, ensembling improved performance under covariate shift. Ensembles were also preferred for the covariate shifts that emerge in motor imagery related electroencephalography (EEG) signals as a user of an EEG-based brain-computer interface experiences changes in attention, fatigue, and other stimuli \cite{Raza_Rathee_Zhou_Cecotti_Prasad_2019}. Combined with ensembling, further interventions can offer benefits, but a straightforward ensemble of a classical model offers a strong baseline.

To complement and explain these empirical findings, we explore the theoretical advantages of ensembling partition-based models. Our studied model isolates the role of ensembling from interventions that attempt to adapt a trained model to the target distribution. Of practical relevance, a modeler may train our model on a mix of source- and target-distributed data without knowing which distribution generated each data point. By contrast, further interventions, including all of those described in the previous paragraph, typically require the modeler to know which of their data points were generated by the target distribution.

\subsection{Notation}

We refer to probability measures by $\mu$ and $\nu$ throughout. Where applicable, $\mu$ refers to the target distribution, which generates the test point $X$ for model evaluation and possibly some of the training data points' feature vectors, while $\nu$ refers to the source distribution, which generates the shifted training points' feature vectors. As such, our source-target pair is $(\nu, \mu)$. Previous works have referred to the source-target pair as $(P, Q)$ \cite{pathak_new_2022} or $(P_X, Q_X)$ \cite{kpotufe_marginal_2021}, to accentuate them being marginal over the covariates. We choose to refer to the target instead by $\mu$ because it is the canonical notation for a measure and we wish to emphasize its importance to the covariate shift problem. The distribution $\mu$ is what the modeler \emph{wants} to learn about, while $\nu$ poses a challenge with which the modeler must contend. 

Our notation for measures aims to minimize confusion with $\mathcal{P}$, by which we refer to a partition of $\R^d$ or of a subset thereof. Uppercase $C$ is an arbitrary cell in a partition. The cell containing the point $x$ is called $Z_x$. In particular, $Z_0$ is the cell containing the origin in $\R^d$. Superscripts on $C$ or $Z_x$ record indices or parameters of the partition to which the cell belongs, not powers. Lowercase $c_\gamma$ is a constant depending on the variable $\gamma$. We use this notation sparingly; in order to avoid confusion with the notation of cell $C$, we prefer to write $a \lesssim b$ to elide a constant $c > 0$ in the inequality $a \leq c \cdot b$. Whenever $\lesssim$ is used, the elided constant's dependence on other variables is noted in the text.

The norm $\|\cdot\|$ is arbitrary unless otherwise noted and $\log$ is the natural logarithm. The notation $\|f\|_\infty$ refers to the maximum absolute value of the real-valued function $f$. By $\mathbf{1}_E$ we refer to the indicator function of event $E$.

\subsection{Related Work on Covariate Shift Discrepancies}

Let $B(x, r)$ refer to the ball of radius $r$ centered at $x$. Previously defined covariate shift discrepancy measures control the growth of the probability ratios $\frac{\mu(B(x, r))}{\nu(B(x, r))}$ as $r$ shrinks. Intuitively, the limit of this ratio, as $r \to \infty$, is the density ratio, which can be used to reweight shifted training data and mitigate the effect of covariate shift \cite{quinonero-candela_dataset_2008,sugiyama_density_2012}. By considering only positive, finite values of $r$, the transfer exponent and other covariate shift discrepancy measures avoid assuming that the density ratio is well-defined \cite{kpotufe_marginal_2021}.

\begin{definition} \label{df:transfer_exponent}
    $\gamma > 0$ is called a \emph{transfer exponent} for the source-target pair of distributions $(\nu, \mu)$ if there is a constant $c_\gamma > 0$ that depends on $\gamma$ such that, for all $r \in (0, \operatorname{diam}(\supp(\mu)))$, 
    \[\sup_{x \in \supp(\mu)} \frac{\mu(B(x, r))}{\nu(B(x, r))} \leq c_\gamma r^{-\gamma}.\]
\end{definition}

Following the introduction of the transfer exponent \cite{kpotufe_marginal_2021}, a refinement was introduced \cite{pathak_new_2022} and named the integrated transfer exponent \cite{Galbraith_Kpotufe_2024}. It controls an integral over the choice of test point $X \sim \mu$, as opposed to the transfer exponent, which controls a supremum over all possible values of $x \in \supp(\mu)$. The integral itself, for fixed $r$, can also be seen as a covariate shift discrepancy; it was so introduced by \cite{pathak_new_2022}, and we refer to it as the integrated $r$-transfer discrepancy.

\begin{definition} \label{df:integrated_transfer_exponent} \label{df:integrated_r-transfer_discrepancy}
    For the source-target pair of distributions $(\nu, \mu)$ and fixed $r > 0$, the \emph{integrated $r$-transfer discrepancy} is defined as \[\E_{X \sim \mu}\left[ \frac{1}{\nu(B(X, r))} \right].\] Moreover, $\gamma > 0$ is called an \emph{integrated transfer exponent} for the source-target pair of distributions $(\nu, \mu)$ if there is a constant $c_\gamma > 0$ that depends on $\gamma$ such that, for all $r > 0$,
    \begin{equation}
        \E_{X \sim \mu}\left[ \frac{1}{\nu(B(X, r))} \right] \leq c_\gamma r^{-\gamma}. \label{ln:integrated_transfer_exponent}
    \end{equation}
\end{definition}
The expression of the integrated $r$-transfer discrepancy emerges when analyzing the variance of the Nadaraya-Watson kernel estimator with the top-hat kernel $k(x, x') = \mathbf{1}_{\|x-x'\| \leq r}$ under covariate shift. The integrated transfer exponent controls the convergence rate as $n$ grows and $r$ correspondingly shrinks, per Theorem 1 in \cite{pathak_new_2022}. 

A closely related discrepancy emerges from analyzing the variance of a dyadic classification tree \cite{Galbraith_Kpotufe_2024}. The local averaging in a dyadic tree is performed over a leaf, which is a cell in the dyadic partition of feature space, so the cells of such partitions replace $B(x, r)$ in the definition.
\begin{definition} \label{df:aggregate_transfer_exponent} \label{df:aggregate_r-transfer_discrepancy}
    An \emph{$r$-grid} $\Xi$ is a partition of $\supp(\nu)$ such that each cell $C \in \Xi$ has a point $x_C$ with 
    $B^\infty(x_C, r) \subseteq C \subseteq B^\infty(x_C, 2r)$ 
    where $B^\infty(x, r)$ is the $\ell_\infty$-norm ball with center $x$ and radius $r$.
    For the source-target pair of distributions $(\nu, \mu)$ and fixed $r > 0$, the \emph{aggregate $r$-transfer discrepancy} is defined as the supremum of \[\sum_{C \in \Xi} \frac{\mu(C)}{\nu(C)}\] over all $r$-grids $\Xi$. Moreover, $\gamma > 0$ is an \emph{aggregate transfer exponent} for the source-target pair of measures $(\nu, \mu)$ if there exists a constant $c_\gamma > 0$ such that, for all $r > 0$ and $r$-grids $\Xi$,
    \begin{equation}
        \sum_{C \in \Xi} \frac{\mu(C)}{\nu(C)} \leq c_\gamma r^{-\gamma}. \label{ln:aggregate_transfer exponent}
    \end{equation}
\end{definition}

Proposition 9 in \cite{Galbraith_Kpotufe_2024} proves that, at least for source-target pairs of doubling measures, the infimum integrated and aggregate transfer exponents coincide.

\subsection{Background on Tessellations and Partitions}

Before introducing partition-based covariate shift discrepancy measures, we must define the class of partitions we study. Let $\mathcal{P}$ be a partition of $\R^d$, meaning a collection of pairwise disjoint sets whose union is $\R^d$. We will generally restrict this partition to $\supp(\nu)$, the set of all possible data or evaluation points. We denote by $Z_x$ the cell of the partition that contains a point $x$ in $\R^d$.

We restrict our attention to deterministic and randomly generated partitions such that the closures of all the cells $\{\operatorname{cl}(C): C \in \mathcal{P}\}$ form a mosaic. A \emph{mosaic} in $\R^d$ is defined as a collection of bounded, convex bodies whose interiors are pairwise disjoint and whose union is $\R^d$ \cite{oreilly_stochastic_2022}. For example, unit $d$-dimensional cubes tile $\R^d$, forming a mosaic \[\{[n_1, n_1 + 1] \times [n_2, n_2 + 1] \times \cdots \times [n_d, n_d + 1] : n_k \in \mathbb{Z}\}.\] To obtain a partition from a mosaic, each boundary between interiors must be assigned to one of the bodies it is contained in. For the grid, this can be accomplished by assigning each face shared between multiple cubes to the uppermost cube in all directions, yielding the partition \[\{[n_1, n_1 + 1) \times [n_2, n_2 + 1) \times \cdots \times [n_d, n_d + 1) : n_k \in \mathbb{Z}\}.\] This and a few other hypotheses about the partition are summarized below.

\begin{hypothesis}[Regarding the Partition] \label{as:partition}
    The closures of the cells $\{\operatorname{cl}(C): C \in \mathcal{P}\}$ form a mosaic in $\R^d$, implying the interior $\operatorname{int}(C)$ is convex for all $C \in \mathcal{P}$. Also, $\mathcal{P}$ is locally finite, meaning for any compact set $K$, the set $\{C \cap K: C \in \mathcal{P}\}$ has finite cardinality. Lastly, $\mathcal{P}$ is generated independently of the test point $X$.
\end{hypothesis}

The union of boundaries of a mosaic is called a \emph{tessellation} \cite{oreilly_stochastic_2022}. The boundaries are less of a concern if the mosaic is determined by a random process with a property called \emph{stationarity}. 

\begin{definition}\label{df:stationarity}
    A random tessellation $\mathcal{T}$ is \emph{stationary} if its distribution is invariant under translation, meaning that for all $x \in \R^d$, the distribution of $\mathcal{T} + x$ is the same as that of $\mathcal{T}$.
\end{definition}

A tessellation has Lesbesgue measure of zero, but it could have nonzero $\mu$- or $\nu$-mass if these measures are not absolutely continuous with respect to Lesbesgue measure. However, if the tessellation is randomly generated and stationary, any fixed point $x$ will almost surely (with respect to the tessellation) fall in the interior of one of the cells, so we can evaluate distributional quantities, like the estimator's expected prediction error of a randomly generated (by $\mu$) test point $X$, without concern for the case that $X$ is contained in the tessellation (as opposed to the interior of a partition cell).

\subsection{Organization}

\Cref{sc:discrepancy_measures} defines covariate shift discrepancies for ensembles of partitions. These partition-based discrepancies have useful theoretical properties (e.g., \cref{pr:integrated_equals_aggregate_P-transfer_discrepancy}) and indicate the utility of ensembling (see \cref{lm:discrepancy_decreases_in_M}). \Cref{sc:general_risk_bounds} provides a risk bound for an ensembled partition-based estimator under covariate shift, demonstrating that our partition-based discrepancy controls the variance. \Cref{sc:STIT_risk_bounds} specializes the risk bound to STIT tessellations and, after presenting a novel bound on the bias (\cref{lm:STIT_bias_bound}), provides a convergence rate for the risk of a STIT kernel estimator. Finally, \Cref{sc:example_shift} evaluates the partition-based discrepancy for a frequently-studied class of severe covariate shifts and demonstrates that the STIT kernel estimator achieves minimax-optimal convergence rates with a sufficiently large ensemble. Three longer proofs are deferred to the appendix.

\section{Partition-based Covariate Shift Discrepancies} \label{sc:discrepancy_measures}

Local averaging estimators can remain consistent under even fairly severe covariate shifts because the shifted training data near a test point have labels similar, on average, to the value of the shifted regression function at the test point, which is the same as the regression function for the test distribution. Covariate shift may erode the estimator's convergence rate (possibly so dramatically that it renders the estimator inconsistent without unshifted data \cite{kpotufe_marginal_2021}) by reducing the speed at which the training set accumulates points near the test point and hence increasing the model's variance; this occurs if the neighborhood around the test point has less probability mass under the shifted (marginal) distribution $\nu$ than the target (marginal) distribution $\mu$. Nevertheless, so long as some distribution generating training data has positive probability on a neighborhood of the test point, eventually enough training data will accumulate near the test point that its partition cell can be refined (or kernel bandwidth shrunk) to reduce the bias 
while controlling the variance.

Quantitatively, transfer exponents appear in place of (or in addition to) the dimension $d$ in the minimax-optimal convergence rates \cite{kpotufe_marginal_2021,pathak_new_2022}. If severe enough to affect minimax rates, a covariate shift has a transfer exponent large enough that $d$ is replaced by a larger number. This makes the learning problem as ``hard'' (i.e., having a minimax rate as slow) as a problem in higher dimensions than $d$.

To define a transfer exponent and reveal it in a model's convergence rate, we require a one-dimensional complexity parameter for the model. For instance, the top-hat kernel bandwidth is a complexity parameter that allows for studying the top-hat kernel estimator, a single learner. However, a general ensemble of partitions may not have a single complexity parameter, so we begin by defining ``transfer discrepancies'' that allow us to derive risk bounds without reference to a model complexity parameter. In \Cref{sc:STIT_risk_bounds}, we study a random partition with a one-dimensional complexity parameter and define a partition-based transfer exponent.

We now define partition-based transfer discrepancies. In anticipation of ensembling random partition-based estimators, we state these definitions with an ensemble size parameter $M$, which is always a positive integer. Partitions $\left\{\mathcal{P}^{(m)}\right\}_{m=1}^M$ are 
identically distributed as $\mathcal{P}$. If $\mathcal{P}$ is deterministic (i.e., a point-mass in the space of partitions), it follows that the realizations $\left\{\mathcal{P}^{(m)}\right\}_{m=1}^M$ are identical. We use the notation $Z_x^m$ to refer to the cell in $\mathcal{P}^{(m)}$ that contains the point $x$.

\begin{definition} \label{df:integrated_P-transfer_discrepancy}
    For a partition $\mathcal{P}$, an ensemble size $M \geq 1$, and a source-target pair of distributions $(\nu, \mu)$, the \emph{integrated $\mathcal{P}^M$-transfer discrepancy} is defined as \[\E\left[ \frac{1}{\nu\left(\bigcup_{m=1}^M Z_X^m\right)} \right],\] where $X$ is distributed as $\mu$. If $\mathcal{P}$ is random, the expectation is taken over the 
    partitions, too.
\end{definition}

In contrast to the integrated $r$-transfer discrepancy (\cref{df:integrated_r-transfer_discrepancy}), the integrated $\mathcal{P}^M$-transfer discrepancy replaces $B(X, r)$ with the union $\bigcup_{m=1}^M Z_X^m$.

The aggregate $\mathcal{P}^M$-transfer discrepancy handles the ensemble intricately.

\begin{definition} \label{df:aggregate_P-transfer_discrepancy}
    For a partition $\mathcal{P}$, an ensemble size $M \geq 1$, and a source-target pair of distributions $(\nu, \mu)$, the \emph{aggregate $\mathcal{P}^M$-transfer discrepancy} is defined as 
    \[\E\left[ \sum_{\substack{{(C^{(m)})_{m=1}^M} \in \prod_{m=1}^M \mathcal{P}^{(m)}: \\ \bigcap_{m=1}^M C^{(m)} \cap \supp(\mu) \ne \varnothing}} \; \frac{\mu\left(\bigcap_{m=1}^M C^{(m)}\right)}{\nu\left(\bigcup_{m=1}^M C^{(m)}\right)} \right],\]
    where the sum is taken over length-$M$ tuples $(C^{(m)})_{m=1}^M$ such that $C^{(m)} \in \mathcal{P}^{(m)}$ for each $m$ and also $\bigcap_{m=1}^M C^{(m)}$ has nonempty intersection with $\supp(\mu)$. The expectation is taken over the realizations of $\mathcal{P}$.
\end{definition}

The aggregate $\mathcal{P}^1$-transfer discrepancy can be directly compared to the aggregate $r$-transfer discrepancy (\cref{df:aggregate_r-transfer_discrepancy}) when $\mathcal{P}$ is almost surely a grid. Since the aggregate $\mathcal{P}^1$-transfer discrepancy is an average, rather than a supremum, it may be smaller. 

\begin{remark} \label{rk:grid_discrepancy}
    Adopt Hypothesis \ref{as:partition}. Suppose $\mathcal{P}$ is almost surely an $r$-grid. Then, the aggregate $\mathcal{P}^1$-transfer discrepancy is no greater than the aggregate $r$-transfer discrepancy. More precisely, if $\mathcal{P}$ is an $r$-grid with probability $1$, then we can compute the aggregate $\mathcal{P}^1$-transfer discrepancy by integrating over this probability-$1$ region only:
    \[ \Prob[\mathcal{P} \text{ is an $r$-grid}] = 1 \implies \E\left[  %
    \sum_{\substack{C \in \mathcal{P}: \\ C \cap \supp(\mu) \ne \varnothing}} \frac{\mu\left(C\right)}{\nu\left(C\right)} \right] \leq \sup_{\Xi \text{ is an $r$-grid}} \sum_{\substack{C \in \Xi: \\ C \cap \supp(\mu) \ne \varnothing}} \frac{\mu\left(C\right)}{\nu\left(C\right)}. \]
\end{remark}

Under certain conditions, it has been shown that the integrated and aggregate transfer exponents are equal \cite{Galbraith_Kpotufe_2024}, but the relation is not known in general. Here we establish equality between the integrated $\mathcal{P}^M$-transfer discrepancy and the aggregate $\mathcal{P}^M$-transfer discrepancy.\footnote{Because integrated and aggregate $\mathcal{P}^M$-transfer discrepancies coincide, we simplify our subsequent discussion by referring only to the integrated $\mathcal{P}^M$-transfer discrepancy by name and rewriting it as the aggregate $\mathcal{P}^M$-transfer discrepancy whenever useful.}

\begin{proposition} \label{pr:integrated_equals_aggregate_P-transfer_discrepancy}
    Adopt Hypothesis \ref{as:partition}. For partition $\mathcal{P}$ and ensemble size $M$, the integrated $\mathcal{P}^M$-transfer discrepancy and aggregate $\mathcal{P}^M$-transfer discrepancy are equal.
\end{proposition}

\begin{proof}
    Consider a fixed $x \in \supp(\mu)$. Each partition $\mathcal{P}^{(m)}$ has exactly one cell $Z_x^m$ that contains $x$, so the Cartesian product $\prod_{m=1}^M \mathcal{P}^{(m)}$ has exactly one tuple $(C^{(m)})_{m=1}^M$ such that $x \in \bigcap_{m=1}^M C^{(m)}$. 
    Moreover, $x$ is in the intersection $\bigcap_{m=1}^M C^{(m)} \cap \supp(\mu)$, hence this intersection is not empty, so the tuple is a member of the subset of the Cartesian product $\prod_{m=1}^M \mathcal{P}^{(m)}$ restricted to $\bigcap_{m=1}^M C^{(m)} \cap \supp(\mu) \ne \varnothing$. Summing an indicator over all such tuples,
    \[\frac{1}{\nu\left(\bigcup_{m=1}^M Z_x^m\right)} = \sum_{\substack{{(C^{(m)})_{m=1}^M} \in \prod_{m=1}^M \mathcal{P}^{(m)}: \\ \bigcap_{m=1}^M C^{(m)} \cap \supp(\mu) \ne \varnothing}} \; \frac{\mathbf{1}_{x \in \bigcap_{m=1}^M C^{(m)}}}{\nu\left(\bigcup_{m=1}^M C^{(m)}\right)}.\]
    The sum above is finite because each partition $\mathcal{P}^{(m)}$ is locally finite and so has finitely many cells with non-empty intersection of the compact set $\supp(\mu)$. Because $X$ is realized independently from the partitions, we can apply linearity of expectation to the integrated $\mathcal{P}^M$-transfer discrepancy, recovering the aggregate $\mathcal{P}^M$-transfer discrepancy as claimed:
    \begin{align*}
        \E\left[ \frac{1}{\nu\left(\bigcup_{m=1}^M Z_X^m\right)} \right] & = \E\left[ \sum_{\substack{{(C^{(m)})_{m=1}^M} \in \prod_{m=1}^M \mathcal{P}^{(m)}: \\ \bigcap_{m=1}^M C^{(m)} \cap \supp(\mu) \ne \varnothing}} \; \frac{\mathbf{1}_{X \in \bigcap_{m=1}^M C^{(m)}}}{\nu\left(\bigcup_{m=1}^M C^{(m)}\right)} \right] \\
        & = \E_{\left\{\mathcal{P}^{(m)}\right\}_{m=1}^M} \left[ \sum_{\substack{{(C^{(m)})_{m=1}^M} \in \prod_{m=1}^M \mathcal{P}^{(m)}: \\ \bigcap_{m=1}^M C^{(m)} \cap \supp(\mu) \ne \varnothing}} \; \frac{ \E_X\left[ \mathbf{1}_{X \in \bigcap_{m=1}^M C^{(m)}} \right] }{\nu\left(\bigcup_{m=1}^M C^{(m)}\right)} \right] \\
        & = \E\left[ \sum_{\substack{{(C^{(m)})_{m=1}^M} \in \prod_{m=1}^M \mathcal{P}^{(m)}: \\ \bigcap_{m=1}^M C^{(m)} \cap \supp(\mu) \ne \varnothing}} \; \frac{\mu\left(\bigcap_{m=1}^M C^{(m)}\right)}{\nu\left(\bigcup_{m=1}^M C^{(m)}\right)} \right].
    \end{align*}
\end{proof}

The integrated transfer exponent is a refinement of the transfer exponent because it considers an average test point $x$, rather than the worst case. The aggregate $r$-transfer exponent approximates an average by summing probability ratios of cells in a partition of $\supp(\mu)$, but it considers the worst-case partition. By randomizing the partition, we obtain a notion of an average partition, allowing us to study the average test point and average model.

The form of the integrated $\mathcal{P}^M$-transfer discrepancy hints at the utility of ensembling because the discrepancy (weakly) decreases as $M$ increases.

\begin{lemma} \label{lm:discrepancy_decreases_in_M}
    Adopt Hypothesis \ref{as:partition}. For source-target pair $(\nu, \mu)$ and partition $\mathcal{P}$, the integrated $\mathcal{P}^M$-transfer discrepancy is decreasing in $M$.
\end{lemma}

\begin{proof}
    Condition on the realizations of the partition. By monotonicity of measure, 
    \[\frac{1}{\nu\left(\bigcup_{m=1}^M Z_X^m\right)} \leq \frac{1}{\nu\left(\bigcup_{m=1}^{M-1} Z_X^m\right)}.\]
    The relation still holds when we take the expectation of both sides.
\end{proof}

The partitions in the ensemble only differ if the partition is randomized, so reductions in the integrated $\mathcal{P}^M$-transfer discrepancy due to ensembling are only possible using random partitions. We will show in \Cref{sc:example_shift} that these reductions can be substantial.

\section{Risk Bounds for General Partition-based Kernel Estimators} \label{sc:general_risk_bounds}

Define the partition-based kernel estimator 
\[\hat{f}_{\mathcal{P}, M}(x) = 
\begin{cases}
    \frac{\sum_{m=1}^M \sum_{i=1}^n Y_i \mathbf{1}_{X_i \in Z_x}}{\sum_{m=1}^M \sum_{i=1}^n \mathbf{1}_{X_i \in Z_x}} & \text{ if } \exists \, m, i : X_i \in Z_x^m \\
    0 & \text{ if } \nexists \, m, i : X_i \in Z_x^m 
\end{cases}.
\]

$\hat{f}_{\mathcal{P}, M}$ is a kernel method whose kernel is called the ``kernel based on random forest'' in \cite{Scornet_2016}. We refer to it as a kernel estimator for brevity.\footnote{The name ``kernel estimator'' applies to $\hat{f}_{\mathcal{P}, M}$ in two senses. Firstly, $\hat{f}_{\mathcal{P}, M}$ estimates the regression function $f^*$ using a kernel induced by the random partitions (a ``kernel based on random forest'' in \cite{Scornet_2016}). Secondly, common choices of random partitions, especially those satisfying Hypothesis \ref{as:purely_random}, admit study of the kernel as $M \to \infty$ and reveal that it approaches a closed-form positive definite kernel. For example, limiting STIT kernels with lifetime $\lambda$ (see \Cref{sc:STIT_risk_bounds}) include the Laplace kernel $k(x, x') = \exp(-\lambda\|x-x'\|_1)$ and the so-called exponential kernel $k(x, x') = \exp(-\lambda\|x-x'\|_2)$, depending on their normalized associated zonoid \cite{oreilly_stochastic_2022}.} Note that it also has the form of a random binning features model \cite{rahimi_random_2007} with bins prescribed by the partitions.

The estimate is learned on the pairwise independent dataset $\{(X_i, Y_i)\}_{i=1}^n$. The feature vectors $\{X_i\}_{i=1}^n$ take values in $\R^d$, and the labels $\{Y_i\}_{i=1}^n$ take values in $\R$. The first $n_\nu \geq 0$ data points $\{(X_i, Y_i)\}_{i=1}^{n_\nu}$ are independently and identically distributed as $\nu_\mathrm{joint}$, a probability measure over $\R^d \times \R$. The remaining $n_\mu = n - n_\nu \geq 0$ data points $\{(X_i, Y_i)\}_{i=n_\nu + 1}^n$ are independently and identically distributed as $\mu_\mathrm{joint}$, another probability measure over $\R^d \times \R$.\footnote{We sort the data points by distribution in this way as a matter of notational convenience. The estimators we study do not require that the modeler knows which distribution gave rise to each data point.} Let $\mu$ and $\nu$ refer to the marginal distributions of $\mu_\mathrm{joint}$ and $\nu_\mathrm{joint}$, respectively, over $\R^d$. In other words, the first $n_\nu$ feature vectors $\{X_i\}_{i=1}^{n_\nu}$ are distributed as $\nu$ and the remaining $n_\mu$ feature vectors $\{X_i\}_{i=n_\nu + 1}^n$ are distributed as $\mu$.

Our risk bounds apply to partition-based kernel estimators whose partitions are independent of the training data $\{(X_i, Y_i)\}_{i=1}^n$. Random forests with this property are called \emph{purely} random forests and are a frequent object of study \cite{Klusowski_2021,Scornet_Biau_Vert_2015,Genuer_2012}. 

\begin{hypothesis}[Purely Random Partitioning] \label{as:purely_random}
    The partition $\mathcal{P}$ is independent of the training data $\{(X_i, Y_i)\}_{i=1}^n$. 
\end{hypothesis}

Dependence on the training data is the primary reason the partitions would have dependence on one another. For simplicity, we assume the partitions are independently realized.

\begin{hypothesis}[Independent Partitions] \label{as:independent_partitions}
    $\left\{\mathcal{P}^{(m)}\right\}_{m=1}^M$ are generated independently.
\end{hypothesis}

The supports of $\mu$ and $\nu$, which are closed by definition, must be nested for the transfer exponent to be finite (Example 1 in \cite{kpotufe_marginal_2021}). We also assume they are compact.

\begin{hypothesis}[Nested and Compact Support] \label{as:support}
The source and target supports are nested, i.e.,    $\supp(\mu) \subseteq \supp(\nu)$. Also, $\supp(\nu)$ is bounded. Hence, $\supp(\mu)$ and $\supp(\nu)$ are compact.
\end{hypothesis}

Additionally, we assume the following about the conditional distribution of $Y|X$ when $(X, Y)$ is sampled from either $\mu_\mathrm{joint}$ or $\nu_\mathrm{joint}$. This is the defining hypothesis of covariate shift, which distinguishes it from general distribution shift. It allows us to focus on the discrepancy between $\mu$ and $\nu$ rather than that of $\mu_\mathrm{joint}$ and $\nu_\mathrm{joint}$.

\begin{hypothesis}[Covariate Shift] \label{as:covariate_shift}
    The regression function $f^*: \supp(\nu) \to \R$ defined as %
    \[f^*(x) := \E [Y | X = x]\]
    is identical for $(X, Y) \sim \mu_\mathrm{joint}$ and $(X, Y) \sim \nu_\mathrm{joint}$.
\end{hypothesis}

Consistent with previous theoretical work on nonparametric regression under covariate shift \cite{pathak_new_2022,Galbraith_Kpotufe_2024}, we study H\"older-continuous regression functions that are Lipshitz or rougher.

\begin{hypothesis}[Smoothness of the Regression Function] \label{as:smoothness}
    There is some $\beta \in (0, 1]$ such that $f^*$ is $\beta$-H\"older-continuous with constant $L$ so that for any $x, x' \in \supp(\nu)$,
    \[|f^*(x) - f^*(x')| \leq L \|x - x'\|^\beta.\]
\end{hypothesis}

Finally, we impose a second-moment bound on the noise in the training labels. The noise $\xi_i := Y_i - f^*(X_i)$ may depend on the features $X_i$ so long as the variance is uniformly bounded.

\begin{hypothesis}[Label Noise] \label{as:noise}
    For every index $i \leq n$, there is a random variable $\xi_i$ that may depend only on $X_i$ (and no other random elements) such that 
    \[Y_i = f^*(X_i) + \xi_i.\]
    Moreover, the second moment of the noise is conditionally bounded by $\sigma^2 > 0$. For all $i \leq n$,
    \[ \sup_{x \in \supp(\nu)} \E\left[ \left. \xi_i^2 \right\rvert X_i = x \right] \leq \sigma^2.\]
    Note that this implies $\E[\xi_i | X_i] = 0$ for all $i$.
\end{hypothesis}

We now derive a pointwise risk bound on $\hat{f}_{\mathcal{P}, M}$. 

\begin{theorem} \label{tm:pointwise_risk_bound}
    Adopt Hypotheses \ref{as:partition} and \ref{as:purely_random} to \ref{as:noise}. For fixed $x \in \supp(\mu)$,
    \begin{align*}
        \E[( \hat{f}_{\mathcal{P}, M}(x) - f^*(x) )^2] & \leq L^2 \cdot \E\left[ \max_{m=1}^M \operatorname{diam}(Z_x^m \cap \supp(\nu))^{2\beta} \right] \nonumber \\
        & \hspace{-1cm} + \frac{18 M \sigma^2 + (f^*(x))^2}{n} \cdot \E\left[n \wedge \frac{1}{\frac{n_\nu}{n} \cdot \nu\left( \bigcup_{m=1}^M Z_x^{m} \right) + \frac{n_\mu}{n} \cdot \mu\left( \bigcup_{m=1}^M Z_x^{m} \right)} \right].
    \end{align*}
\end{theorem}

The proof of \cref{tm:pointwise_risk_bound} is provided in \Cref{sc:pointwise_risk_bound_proof}. 
If the partition is stationary, we can take the expectation with respect to a random test point $X \sim \mu$. This yields an integrated risk bound that depends on a truncation of the integrated $\mathcal{P}^M$-transfer discrepancy.

\begin{corollary} \label{cr:integrated_stationary_risk_bound}
    Adopt Hypotheses \ref{as:partition} and \ref{as:purely_random} to \ref{as:noise}, and suppose $\mathcal{P}$ is stationary. Then,
    \begin{align*}
        \E[(\hat{f}_{\mathcal{P}, M}(x) - f^*(x))^2] & \leq L^2 \cdot \E\left[ \max_{m=1}^M \operatorname{diam}(Z_0^m)^{2\beta} \right] \nonumber \\
        & \hspace{-.5cm} + \frac{18 M \sigma^2 + \|f^*\|_\infty^2}{n} \cdot \E\left[n \wedge \frac{1}{\frac{n_\nu}{n} \cdot \nu\left( \bigcup_{m=1}^M Z_X^{m} \right) + \frac{n_\mu}{n} \cdot \mu\left( \bigcup_{m=1}^M Z_X^{m} \right)} \right].
    \end{align*}    
\end{corollary}

\begin{proof}
    Stationarity means that, for any point $x$ and index $m$, the cell $Z_x^m$ is distributed as $Z_0^m + x$. The diameter is a translation-invariant function of the cell, so we can translate $Z_x^m$ back to the origin without changing the diameter, thus obtaining a bound for the bias term that does not depend on $x$:
    \[\E\left[ \max_{m=1}^M \operatorname{diam}(Z_x^m \cap \supp(\nu))^{2\beta} \right] \leq \E\left[ \max_{m=1}^M \operatorname{diam}(Z_x^m)^{2\beta} \right] = \E\left[ \max_{m=1}^M \operatorname{diam}(Z_0^m)^{2\beta} \right].\]
    We combine this observation with \cref{tm:pointwise_risk_bound} and take the supremum of $f^*$ to conclude.
\end{proof}

Observe that the latter term in \cref{cr:integrated_stationary_risk_bound}, which controls the variance, is no greater than the integrated $\mathcal{P}^M$-transfer discrepancy for source target pair $\left( \left( \frac{n_\nu}{n} \cdot \nu + \frac{n_\mu}{n} \cdot \mu \right), \mu \right)$. 

\section{Convergence Rates for STIT Kernel Estimators} \label{sc:STIT_risk_bounds}

For the remainder of the paper, we focus on random partitions induced by independently realized STIT tessellations. We introduce the tessellation, define a specialized transfer exponent, and derive a convergence rate for the associated partition-based kernel estimator.

\subsection{STIT Tessellation} \label{sc:STIT_definition}

For a construction of the STIT tessellation, we refer to \cite{Nagel_Weiss_2003} and \cite{oreilly_stochastic_2022}. To state our results, it suffices to present only the defining parameters and characteristics of the STIT tessellation. STIT (``stable under iteration'') tessellations are a class of stationary random tessellations parameterized by deterministic convex bodies called centered zonoids. The support function of a centered zonoid $\Pi'$ can be expressed as 
\[h(\Pi', v) := \sup_{x \in \Pi'} \langle v, x \rangle = \frac{\lambda}{2} \int_{\mathbb{S}^{d-1}} |\langle u, v \rangle| \; d\phi(u),\] 
where $\phi$ is an even probability measure on the sphere $\mathbb{S}^{d-1}$ and $\lambda$ is a positive real number. We refer to $\lambda$ as the \emph{lifetime} and $\phi$ as the \emph{directional distribution}. The STIT tessellation gets its name from its property of being stable under iteration, from which follows the \emph{scaling property}: the $\frac{1}{\lambda}$ dilation of the tessellation with lifetime $\lambda$ and directional distribution $\phi$ has the same distribution as the tessellation of lifetime $1$ and directional distribution $\phi$. The zonoid $\Pi$ associated to the tessellation of lifetime $1$ is called the \emph{normalized zonoid} associated to all STIT tessellations with its same directional distribution $\phi$.

As such, whenever $\mathcal{P}$ is induced by a STIT tessellation, we can fully describe its distribution by a pair $(\lambda, \Pi)$. The lifetime $\lambda$ captures the complexity, with higher values of $\lambda$ corresponding to finer partitions. The normalized associated zonoid $\Pi$ captures the relative likelihood of cell boundaries oriented in various directions.

\begin{figure}
    \centering
    \includegraphics[width=1\linewidth]{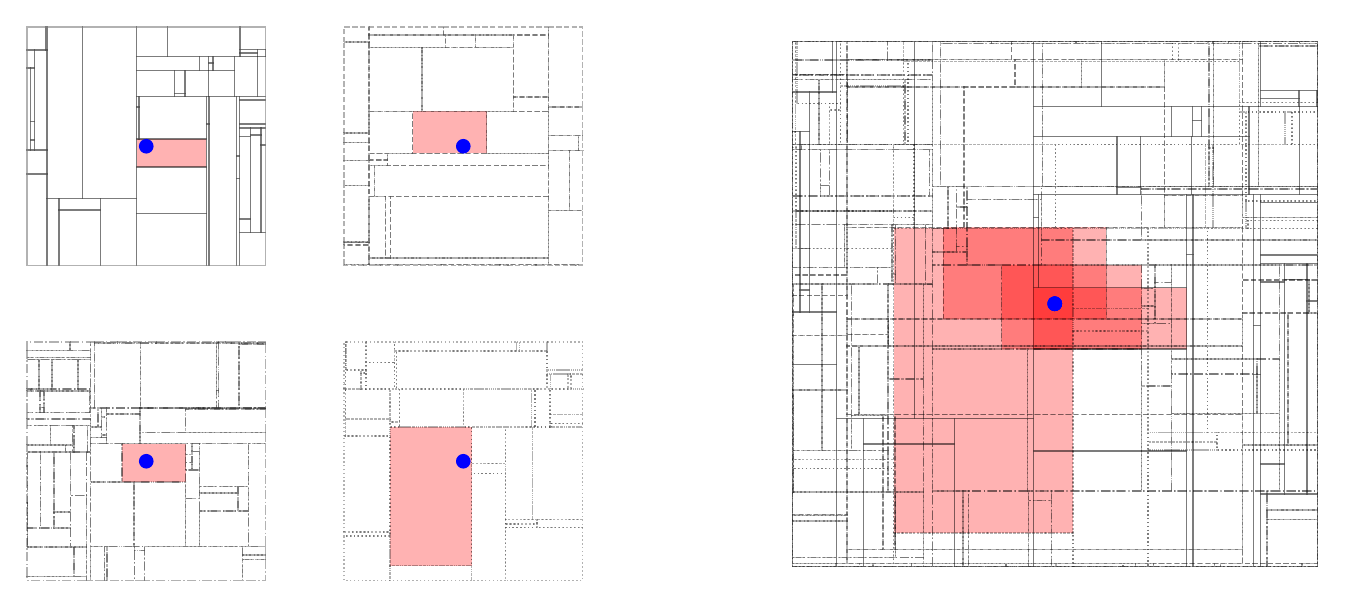}
    \caption{Four realizations of the Mondrian tessellation-induced partition of the square $[0, 1]^2$ are shown individually on the left and superimposed on the right. The test point $x = \left(\frac{1}{2}, \frac{1}{2}\right)$ is marked in blue, and the cell $Z_x$ is shaded in red. The union of the independently realized partitions' cells $Z_x$ is larger than any single partition's, illustrating how ensembling reduces the $\mathcal{P}^M$-transfer discrepancy (\cref{lm:discrepancy_decreases_in_M}).}
    \label{fig:superimposed_Mondrian}
\end{figure}

For example, the normalized zonoid associated to the \emph{Mondrian} tessellation is the $d$-dimensional axis-aligned cube with side length $\frac{d}{2}$, so all cell boundaries in the Mondrian are axis-aligned, and boundaries are equally likely to be aligned to any of the axes. Four realizations of the Mondrian in dimension $d = 2$ (with fixed lifetime $\lambda$) are shown in \cref{fig:superimposed_Mondrian}, both individually and superimposed. For the test point $x = \left(\frac{1}{2}, \frac{1}{2}\right)$, the cell $Z_x$ is highlighted in each. The union of these cells is (stochastically) larger than any single one, which is why ensembling reduces the $\mathcal{P}^M$-transfer discrepancy (see \cref{lm:discrepancy_decreases_in_M}).

\begin{figure}
    \centering
    \includegraphics[width=1\linewidth]{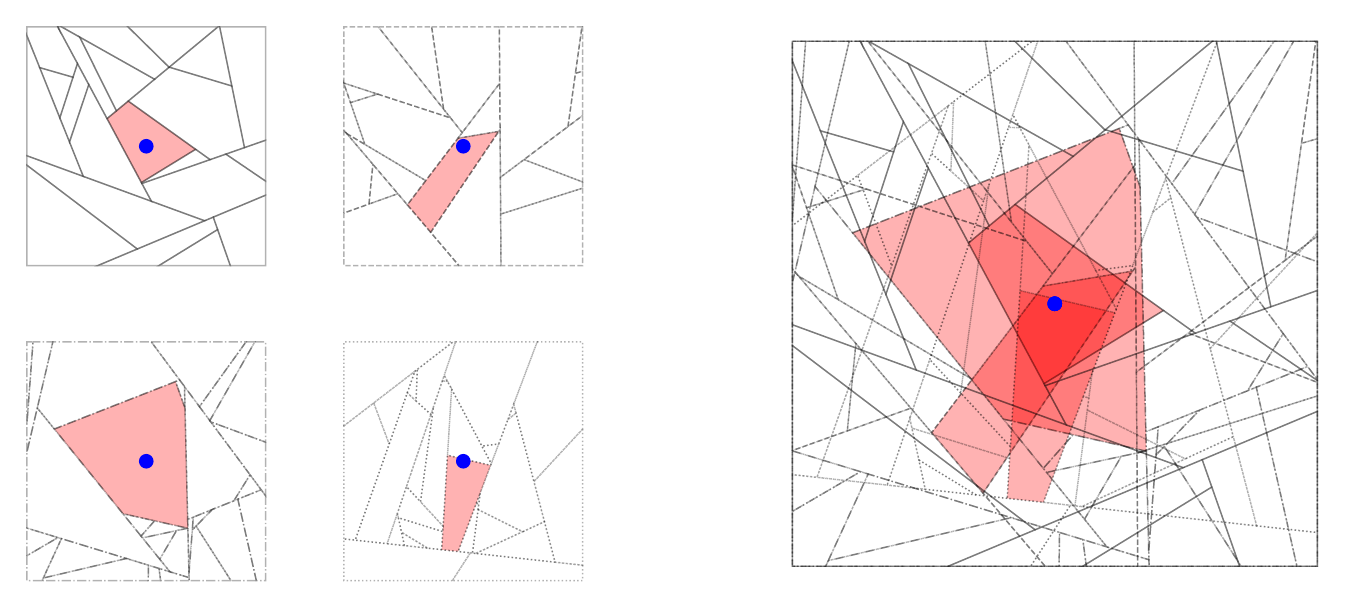}
    \caption{Four realizations of the isotropic STIT tessellation-induced partition of the square $[0, 1]^2$ are shown individually on the left and superimposed on the right. The test point $x = \left(\frac{1}{2}, \frac{1}{2}\right)$ is marked in blue, and the cell $Z_x$ is shaded in red.}
    \label{fig:superimposed_STIT}
\end{figure}

If $\Pi$ is a Euclidean ball, the STIT tessellation is called \emph{isotropic}. \Cref{fig:superimposed_STIT} shows four realizations of the isotropic STIT and their superposition.

The lifetime $\lambda$ is central to the theoretical analysis of STIT tessellation estimators because tuning $\lambda$ allows modelers to balance the bias-variance tradeoff and achieve optimal convergence rates. The directional distribution plays a subtler role, affecting the dependence of risk bounds on $d$ but not convergence rates in $n$. Hence, we use $\lambda$ as a shorthand for the parameterization of a STIT tessellation used in defining estimators or discrepancies. $\mathcal{P}_\lambda$ is our notation for the partition induced by a STIT tessellation parameterized by $(\lambda, \Pi)$, and $Z_x^{m, \lambda}$ refers to the cell in the partition $\mathcal{P}_\lambda^{(m)}$ that contains the point $x$. The partitions $\left\{\mathcal{P}_\lambda^{(m)}\right\}_{m=1}^M$ are always generated independently of one another and so satisfy Hypothesis \ref{as:independent_partitions}.

\begin{remark}
    $\mathcal{P}_\lambda$ satisfies Hypotheses \ref{as:partition} and \ref{as:purely_random} and is stationary. 
\end{remark}

\subsection{STIT Covariate Shift Discrepancy Measures}

For STIT tessellations, we can define transfer exponents by letting the lifetime $\lambda$ take the role of the inverse bandwidth $r^{-1}$. 

\begin{definition} \label{df:P_l-transfer_exponent}
    For ensemble size $M \geq 1$, we call $\gamma \geq 0$ a \emph{$\mathcal{P}^M_\lambda$-transfer exponent} if, for all $\lambda > \operatorname{diam}(\supp(\mu))$, with probability $1$ with respect to the partitions $\left\{\mathcal{P}_\lambda^{(m)}\right\}_{m=1}^M$,
    \[\sup_{x \in \supp(\mu)} \frac{\mu\left( \bigcap_{m=1}^M Z_x^{m, \lambda} \right)}{\nu\left( \bigcup_{m=1}^M Z_x^{m, \lambda} \right)} \lesssim \lambda^\gamma,\]
    where $\lesssim$ elides a constant that depends on $\gamma$, $M$, and $\Pi$.
\end{definition}

\begin{definition} \label{df:integrated_P_l-transfer_exponent}
    For ensemble size $M \geq 1$, we call $\gamma > 0$ an \emph{integrated $\mathcal{P}^M_\lambda$-transfer exponent} if, for all $\lambda > 0$,
    \[\E \left[ \frac{1}{\nu\left(\bigcup_{m=1}^M Z_X^{m, \lambda}\right)} \right] \lesssim \lambda^\gamma,\] 
    where the expectation is taken over the independently realized partitions and $X$, which is distributed as $\mu$. Here, $\lesssim$ elides a constant that depends on $\gamma$, $M$, and $\Pi$.
\end{definition}

The $\mathcal{P}^M_\lambda$-transfer exponent can be seen as dimension-free, whereas the integrated $\mathcal{P}^M_\lambda$-transfer exponent incorporates dimension in the following sense. If $\gamma > 0$ is a $\mathcal{P}^1_\lambda$-transfer exponent, then $\gamma + d_\mu$ is an integrated $\mathcal{P}^M_\lambda$-transfer exponent for any ensemble size $M \geq 1$, where $d_\mu$ is the dimension of $\supp(\mu)$. However, this relationship is loose; there are source-target pairs with integrated $\mathcal{P}^M_\lambda$-transfer exponents less than $\gamma + d_\mu$ as we will show in \Cref{sc:example_shift}. A similar relationship holds between the transfer exponent and integrated transfer exponent and is why the latter is seen as a refinement of the former \cite{Galbraith_Kpotufe_2024,pathak_new_2022}. 

\begin{proposition}
    Suppose $\gamma > 0$ is a $\mathcal{P}^1_\lambda$-transfer exponent for source-target pair $(\nu, \mu)$. Then $\gamma + d_\mu$ is an integrated $\mathcal{P}^M_\lambda$-transfer exponent for any ensemble size $M \geq 1$.
\end{proposition}

\begin{proof}
    Due to \cref{lm:discrepancy_decreases_in_M} and \cref{pr:integrated_equals_aggregate_P-transfer_discrepancy}, 
    \[\E \left[ \frac{1}{\nu\left(\bigcup_{m=1}^M Z_X^{m, \lambda}\right)} \right] \leq \E \left[ \frac{1}{\nu\left(Z_X^{1, \lambda}\right)} \right] = \E \left[ \sum_{\substack{C \in \mathcal{P}_\lambda^{(1)}: \\ C \cap \supp(\mu) \ne \varnothing}} \; \frac{\mu\left(C\right)}{\nu\left(C\right)} \right],\]
    and we need only show that $\gamma + d_\mu$ is an aggregate $\mathcal{P}^1_\lambda$-transfer exponent. 

    For each $C \in \mathcal{P}_\lambda^{(1)}: C \cap \supp(\mu) \ne \varnothing$, there is some $x \in \supp(\mu)$ such that $C = Z_x^{1, \lambda}$. Thus, we can apply the definition of the $\mathcal{P}^1_\lambda$-transfer exponent to bound 
    \begin{align*}
        \E \left[ \sum_{\substack{C \in \mathcal{P}_\lambda^{(1)}: \\ C \cap \supp(\mu) \ne \varnothing}} \; \frac{\mu\left(C\right)}{\nu\left(C\right)} \right] & \lesssim \E \left[ \sum_{\substack{C \in \mathcal{P}_\lambda^{(1)}: \\ C \cap \supp(\mu) \ne \varnothing}} \; \lambda^\gamma \right] \\
        & \lesssim \lambda^\gamma \cdot \E\left[ \# \left\{ C \in \mathcal{P}_\lambda^{(1)}: C \cap \supp(\mu) \ne \varnothing \right\} \right] \\
        & \lesssim \lambda^\gamma \cdot \lambda^{d_\mu}.
    \end{align*}
    The last line above is a consequence of Lemma 4 in \cite{oreilly_minimax_2024}, and the $\lesssim$ elides lower-degree terms in $\lambda$ and coefficients that depend on $\gamma$, $\Pi$, and $d_\mu$ but not $\lambda$.
\end{proof}

\subsection{STIT Kernel Estimator Convergence Rate}

We next specialize our risk bound in \Cref{tm:pointwise_risk_bound} to the STIT kernel estimator and use the integrated $\mathcal{P}^M_\lambda$-transfer exponent to control the variance as $\lambda$ grows with $n$. First, we obtain a bound for the bias of a STIT kernel estimator. This bound grows logarithmically in the ensemble size $M$, so it is suitable for moderate-to-large ensembles but not extremely large ones. The general STIT kernel estimator's bias bound depends on $d$ through the associated zonoid $\Pi$ via a constant $c_{\beta, \Pi}$. To shed light on this dependence on $d$, we separately derive a bias bound for the Mondrian kernel estimator, using the explicit distribution of the zero cell of the Mondrian partition \cite{Mourtada_Gaiffas_Scornet_2020}.

\begin{lemma} \label{lm:STIT_bias_bound}
    There exists $c_{\beta, \Pi} > 0$ depending on $\beta$ and $\Pi$ such that for any $x \in \supp(\mu)$,
    \[\E\left[ \max_{m=1}^M \operatorname{diam}(Z_x^{m, \lambda})^{2\beta} \right] \lesssim \frac{c_{\beta, \Pi}}{\lambda^{2\beta}} \cdot \left(1 \vee \log^{2\beta} (dM)\right).\]
    For Mondrian partitions, with $M \geq 2$,
    \[\E\left[ \max_{m=1}^M \operatorname{diam}(Z_x^{m, \lambda})^{2\beta} \right] \lesssim \frac{d^{\beta}}{\lambda^{2\beta}} \log^{2\beta} M.\]
    In both cases, $\lesssim$ elides a universal constant.
\end{lemma}

Proofs of the two statements in \cref{lm:STIT_bias_bound} are provided in \Cref{sc:STIT_bias_bound_proof} and \Cref{sc:proof_Mondrian_bias_bound}.
Together, \cref{cr:integrated_stationary_risk_bound} and \cref{lm:STIT_bias_bound}, bound the integrated risk of a STIT kernel estimator.

\begin{theorem} \label{tm:integrated_STIT_risk_bound}
    Adopt Hypotheses \ref{as:support} to \ref{as:noise}. Then,
    \begin{align*}
        \E[(\hat{f}_{\mathcal{P}_\lambda, M}(x) - f^*(x))^2] & \lesssim \frac{L^2 c_{\beta, \Pi}}{\lambda^{2\beta}} \cdot \left(1 \vee \log^{2\beta} (dM)\right) \\
        & \hspace{-.5cm} + \frac{M \sigma^2 + \|f^*\|_\infty^2}{n} \cdot \E\left[n \wedge \frac{1}{\frac{n_\nu}{n} \cdot \nu\left( \bigcup_{m=1}^M Z_X^{m, \lambda} \right) + \frac{n_\mu}{n} \cdot \mu\left( \bigcup_{m=1}^M Z_X^{m, \lambda} \right)} \right].
    \end{align*}
    Here, $\lesssim$ elides a universal constant.
\end{theorem}

Now that both the bias and variance are controlled by expressions involving the lifetime parameter $\lambda$, we can tune $\lambda$ to balance the bias and variance, obtaining a convergence rate. Compared to the convergence rate of the STIT tree without covariate shift (Corollary 7 in \cite{oreilly_minimax_2024}), this rate has the integrated $\mathcal{P}^M_\lambda$-transfer exponent in the role of $d$.

\begin{corollary} \label{cr:STIT_convergence_rate}
    If $\gamma \geq d$ is an integrated $\mathcal{P}^M_\lambda$-transfer exponent for the source-target pair $\left(\left(\frac{n_\nu}{n} \cdot \nu + \frac{n_\mu}{n} \cdot \mu \right), \mu \right)$, then tuning $\lambda \asymp n^{\frac{1}{2\beta + \gamma}}$ causes $\hat{f}_{\mathcal{P}_\lambda, M}$ to achieve the following convergence rate in $n$:
    \[\E[(\hat{f}_{\mathcal{P}_\lambda, M}(x) - f^*(x))^2] \lesssim \frac{1}{\lambda^{2\beta}} + \frac{\lambda^\gamma}{n} \asymp n^{-\frac{2\beta}{2\beta + \gamma}}.\]
    Here, $\lesssim$ elides constants depending on $\gamma$, $L$, $\beta$, $d$, $\Pi$, $M$, $\sigma^2$, and $\|f^*\|_\infty$ but not $\lambda$.
\end{corollary}

This convergence rate is minimax-optimal for $n_\nu = n$ and $d=1$. A matching lower bound is provided in Theorem 2 of \cite{pathak_new_2022}, wherein the authors say the proof can be extended to $d > 1$.

\begin{remark}
    In our convergence rates, the integrated $\mathcal{P}^M_\lambda$-transfer exponent $\gamma$ takes the role of $d$. If $\gamma < d$, that would suggest the covariate-shifted training data confers rate improvements over unshifted data. Intuitively, this could occur if the source $\nu$ were closer to the uniform distribution than the target $\mu$. For further discussion of the integrated transfer exponent being strictly less than $d$, see Sections 2.3 and 2.4 in \cite{pathak_new_2022}. 
\end{remark}

\section{STIT Kernel Estimator on Example Covariate Shift} \label{sc:example_shift}

\begin{figure}
    \centering
    \includegraphics[]{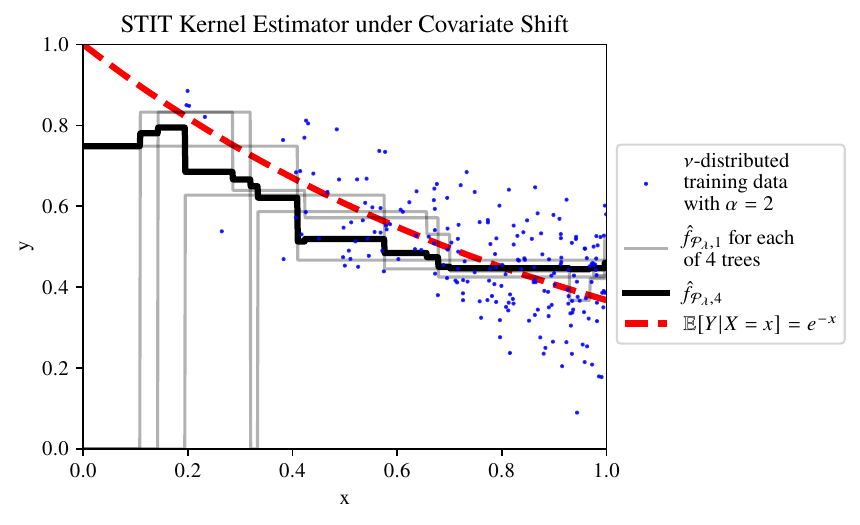}
    \caption{The STIT kernel estimator is used to fit an exponential regression function under covariate shift of severity $\alpha = 2$. The covariate shift results in insufficient data near $0$, so individual trees (i.e., single learners) have high variance. Ensembling smooths the estimate, which improves its performance, especially near $0$.}
    \label{fig:STIT_kernel_estimator}
\end{figure}

In this section, we demonstrate that ensemble size can significantly affect the integrated $\mathcal{P}^M_\lambda$-transfer exponent. We study a class of covariate shifts that have previously been used to demonstrate properties of covariate shift discrepancies, such as the separation between the transfer exponent and integrated transfer exponent \cite{pathak_new_2022,Galbraith_Kpotufe_2024}. For covariate shifts in this class, $\mu$ is the uniform measure over the unit cube and $\nu$ has a polynomial density. We restrict our focus to $d=1$, where the only STIT tessellation is the Mondrian and the distribution of $\bigcup_{m=1}^M Z_X^{m, \lambda}$ can be expressed directly. As in other works, we suppose all training data is $\nu$-distributed, so $n_\nu = n$ \cite{pathak_new_2022,Galbraith_Kpotufe_2024}. This covariate shift, and the performance of the STIT kernel estimator on it, are illustrated in \cref{fig:STIT_kernel_estimator}. 

\begin{hypothesis}[Polynomial Density] \label{as:polynomial_density}
    $\mu = \operatorname{Unif}([0, 1])$, and there is an integer $\alpha > 1$ such that $\nu$ has density $(\alpha + 1) x^\alpha \mathbf{1}_{x \in [0, 1]}$. 
\end{hypothesis}

\begin{remark}
   Under Hypothesis~\ref{as:polynomial_density}, the infimum transfer exponent is $\alpha + 1$, while the infimum integrated transfer exponent is $\alpha$ \cite{pathak_new_2022,Galbraith_Kpotufe_2024}.
\end{remark}

We show that the integrated $\mathcal{P}^M_\lambda$-transfer discrepancy for the Mondrian kernel estimator is infinite for $M \leq \alpha$ yet the Mondrian kernel estimator achieves the minimax-optimal convergence rates for $M \geq \alpha + 2$. 

\begin{proposition} \label{pr:insufficient_ensemble_size}
    Adopt Hypothesis~\ref{as:polynomial_density}. If $M \leq \alpha$, the integrated $\mathcal{P}^M_\lambda$-transfer discrepancy is infinite.
\end{proposition}

The proof is provided in \Cref{sc:proofs_example_shift}. It uses the distribution of the zero cell of the Mondrian tessellation \cite{Mourtada_Gaiffas_Scornet_2020} to relate the integrated $\mathcal{P}^M_\lambda$-transfer discrepancy to an inverse moment of a Gamma-distributed random variable with shape parameter $M$. When the shape parameter is small, the Gamma random variable has very few finite inverse moments.

A larger ensemble size allows us to better control the variance of the estimator. While the stochastic nature of the partition complicates efforts to bound the integrated $\mathcal{P}^M_\lambda$-transfer exponent itself, we can nevertheless use \cref{tm:integrated_STIT_risk_bound} and \cref{cr:STIT_convergence_rate} to guarantee the Mondrian kernel estimator with sufficient $M$ achieves minimax-optimal convergence rates.

\begin{proposition} \label{pr:sufficient_ensemble_size}
    Adopt Hypothesis~\ref{as:polynomial_density}. If $M \geq \alpha + 2$, the Mondrian kernel estimator achieves the minimax-optimal convergence rate. Namely,
    \begin{align*}
        \E[(\hat{f}_{\mathcal{P}_\lambda, M}(X) - f^*(X))^2] & \lesssim \frac{\log^{2\beta} M}{\lambda^{2\beta}} + \frac{\lambda^{\alpha}}{n} \left( \frac{M^2}{\left( 1 - \frac{\alpha+1}{M} \right)^{\alpha+1}} + \frac{M}{\left( 1 - \frac{\alpha}{M} \right)^\alpha} \right) + \frac{M^3 e^{-\lambda}}{n \lambda} + M^3 e^{-\frac{\lambda}{2}},
    \end{align*}
    where $\lesssim$ elides coefficients that depend on $L$, $\sigma^2$, and $\|f^*\|_\infty$.
    
    We tune $\lambda \asymp n^{-\frac{1}{2\beta + \alpha}}$ and, so long as $M$ is not too large (i.e., no more than polynomial in $\lambda$), then for sufficient $n$, the risk has the minimax-optimal \cite{pathak_new_2022} rate of $n^{-\frac{2\beta}{2\beta + \alpha}}$.
\end{proposition}

The proof is provided in \Cref{sc:sufficient_ensemble_size_proof}. 
In effect, we prove in \cref{pr:sufficient_ensemble_size} that for sufficient ensemble size, the integrated $\mathcal{P}^M_\lambda$-transfer exponent is $\alpha$ with high probability. 
The truncation at $n$ in $\E\left[n \wedge \frac{1}{\nu\left(\bigcup_{m=1}^M Z_X^{m, \lambda}\right)}\right]$ allows us to control the risk due to the low-probability event. This principle is generalized in the following corollary to \cref{tm:integrated_STIT_risk_bound}.

\begin{corollary} \label{cr:high_probability_STIT_convergence_rate}
    Adopt Hypotheses \ref{as:support} to \ref{as:noise}. Suppose there is a number $\gamma \geq d$ and an event $E$ in the joint probability space of $X \sim \mu$ and $\left\{ \mathcal{P}_\lambda^{(m)} \right\}_m$ such that 
    \[\E\left[ \frac{\mathbf{1}_E}{\nu\left(\bigcup_{m=1}^M Z_X^{m, \lambda}\right)} \right] \lesssim \lambda^\gamma,\]
    where $\lesssim$ elides coefficients depending on $\gamma$ and $M$ but not $\lambda$.
    Also, suppose $E$ has probability at least $1 - \exp\left(-\frac{c\lambda}{\log M}\right)$ for some universal constant $c > 0$.
    Finally, suppose $\lambda$ is tuned to $\lambda \asymp n^{\frac{1}{2\beta + \gamma}}$. Then, the conclusion of \cref{cr:STIT_convergence_rate} holds with  $\gamma$ in the role of the integrated $\mathcal{P}^M_\lambda$-transfer exponent. Namely, the STIT kernel estimator $\hat{f}_{\mathcal{P}_\lambda, M}$ has the following integrated risk bound:
    \[\E[(\hat{f}_{\mathcal{P}_\lambda, M}(x) - f^*(x))^2] \lesssim \frac{1}{\lambda^{2\beta}} + \frac{\lambda^\gamma}{n} + \exp\left(-c\lambda\right) \asymp n^{-\frac{2\beta}{2\beta + \gamma}}.\]
    Here, $\lesssim$ elides constants depending on $\gamma$, $L$, $\beta$, $d$, $\Pi$, $M$, $\sigma^2$, and $\|f^*\|_\infty$ but not $\lambda$.
\end{corollary}

\section{Conclusion}

In this paper, we prove theoretical guarantees for partition-based kernel estimators under covariate shift, demonstrating the beneficial effect of ensembling. We begin by modifying previously introduced covariate shift measures \cite{kpotufe_marginal_2021,pathak_new_2022,Galbraith_Kpotufe_2024} to control the variance of partition-based kernel estimators. Our pointwise risk bound applies to general partitions. We then derive an integrated risk bound specifically for STIT tessellations under covariate shift, whose complexity is prescribed by the lifetime parameter $\lambda$. Lastly, we analyze a class of well-studied severe covariate shifts, revealing that ensemble size must increase with shift severity to control the partition-based discrepancy measure (with sufficiently high probability).  

Our conclusions about the importance of ensembling parallel recent insights about Mondrian forests, where ensembling counteracts the small-cell phenomenon and enables statistical inference \cite{Cattaneo_Klusowski_Underwood_2025}. We find that ensembling likewise helps under covariate shift to control variance. 

Empirical evidence suggests ensembles help to transfer learning from one distribution to another, even under more general distribution shifts \cite{Clark_Yatskar_Zettlemoyer_2019}. To illuminate and explain this phenomenon theoretically, we envision extending and evaluating discrepancy measures for posterior drift \cite{Cai_Wei_2021,Cai_Pu_2024} to ensemble models. Recent work suggests ensembling might even control variance well enough to explain the success of interpolating models \cite{Wu_Sun_2025}, another fruitful avenue for continued theoretical work to illuminate and explain the empirical success of ensembles.

\section{Proof of \Cref{tm:pointwise_risk_bound}} \label{sc:pointwise_risk_bound_proof}

\begin{proof}[Proof of \cref{tm:pointwise_risk_bound}]
    Conditioning on the training features $\{X_i\}_{i=1}^n$ and the partitions $\left\{\mathcal{P}^{(m)}\right\}_{m=1}^M$, we compute the conditional expectation of the kernel estimator:  %
    \begin{align*}
        \Bar{f}_{\mathcal{P}, M}(x) &:= \E\left[ \left. \hat{f}_{\mathcal{P}, M}(x) \right\rvert \{X_i\}_{i=1}^n, \left\{\mathcal{P}^{(m)}\right\}_{m=1}^M \right] \\
        & = \mathbf{1}_{\exists i, m: X_i \in Z_x^{m}} \frac{\frac{1}{M} \sum_{m=1}^M \left( \sum_{i=1}^{n_\nu} f^*_\nu(X_i) \mathbf{1}_{X_i \in Z_x^{m}} + \sum_{i=n_\nu + 1}^n f^*_\mu(X_i) \mathbf{1}_{X_i \in Z_x^{m}} \right)}{\frac{1}{M} \sum_{m=1}^M \sum_{i=1}^n \mathbf{1}_{X_i \in Z_x^{m}}}.
    \end{align*}
    
    Conditioned on the training features $\{X_i\}_{i=1}^n$ and the partitions $\left\{\mathcal{P}^{(m)}\right\}_{m=1}^M$, we separate out the case in which there is no local data and perform a bias-variance decomposition:
    
    \begin{align}
        \E\left[\left( \hat{f}_{\mathcal{P}, M}(x) - f^*(x) \right)^2\right] & = \E\left[\left( \left( \hat{f}_{\mathcal{P}, M}(x) - \Bar{f}_{\mathcal{P}, M}(x) \right) + \left( \Bar{f}_{\mathcal{P}, M}(x) - f^*(x) \right) \right)^2 \mathbf{1}_{\exists i, m : X_i \in Z_x^{m}}\right] \nonumber \\
        & \hspace{.5cm} + \E\left[ \left( \hat{f}_{\mathcal{P}, M}(x) - f^*(x) \right)^2 \mathbf{1}_{\nexists i, m : X_i \in Z_x^{m}} \right] \nonumber \\
        & = \E\left[\left( \hat{f}_{\mathcal{P}, M}(x) - \Bar{f}_{\mathcal{P}, M}(x) \right)^2 \mathbf{1}_{\exists i, m : X_i \in Z_x^{m}}\right] \label{ln:bias_variance_decomposition}  \\
        & \hspace{.5cm} + \E\left[\left( \Bar{f}_{\mathcal{P}, M}(x) - f^*(x) \right)^2 \mathbf{1}_{\exists i, m : X_i \in Z_x^{m}}\right] \nonumber \\
        & \hspace{.5cm} + \E\left[ (f^*(x))^2 \cdot \Prob\left[ \nexists i, m: X_i \in Z_x^{m} \left\rvert \left\{\mathcal{P}^{(m)}\right\}_{m=1}^M \right. \right] \right]. \nonumber 
    \end{align}
    The cross-terms cancel due to a conditioning argument, which hinges on the definition of $\Bar{f}_{\mathcal{P}, M}(x)$ and the observation that neither it nor $f^*(x)$ depend on $\{\xi_i\}_i$. Namely,
    \begin{align*}
        & \E\left[\left( \hat{f}_{\mathcal{P}, M}(x) - \Bar{f}_{\mathcal{P}, M}(x) \right) \left( \Bar{f}_{\mathcal{P}, M}(x) - f^*(x) \right) \mathbf{1}_{\exists i, m : X_i \in Z_x^{m}}\right] \\
        & \hspace{.2cm} = \E\left[ \left. \E\left[\left( \hat{f}_{\mathcal{P}, M}(x) - \Bar{f}_{\mathcal{P}, M}(x) \right) \left( \Bar{f}_{\mathcal{P}, M}(x) - f^*(x) \right) \mathbf{1}_{\exists i, m : X_i \in Z_x^{m}} \right\rvert \{X_i\}_{i=1}^n, \left\{\mathcal{P}^{(m)}\right\}_{m=1}^M \right] \right] \\
        & \hspace{.2cm} = \E\left[ \left. \left( \E\left[ \hat{f}_{\mathcal{P}, M}(x) \right\rvert \{X_i\}_{i=1}^n, \left\{\mathcal{P}^{(m)}\right\}_{m=1}^M \right] - \Bar{f}_{\mathcal{P}, M}(x) \right) \left( \Bar{f}_{\mathcal{P}, M}(x) - f^*(x) \right) \mathbf{1}_{\exists i, m : X_i \in Z_x^{m}} \right] = 0.
    \end{align*}
    
    We bound the pointwise bias by a convex combination of the diameters of the cells $Z_x^{m}$, provided at least one of these cells contains training points. Using Jensen's inequality (over a finite average) and H\"older continuity, 
    
    \begin{align}
        \left( \Bar{f}_{\mathcal{P}, M}(x) - f^*(x) \right)^2 \mathbf{1}_{\exists i, m : X_i \in Z_x^{m}} & = \left( \frac{\sum_{i=1}^n (f^*(X_i) - f^*(x)) \left( \sum_{m=1}^M \mathbf{1}_{X_i \in Z_x^m}\right) }{\sum_{i=1}^n \left( \sum_{m=1}^M \mathbf{1}_{X_i \in Z_x^m}\right) } \right)^2 \nonumber \\
        & \hspace{-.5cm} \leq \frac{\sum_{i=1}^n (f^*(X_i) - f^*(x))^2 \left( \sum_{m=1}^M \mathbf{1}_{X_i \in Z_x^m}\right) }{\sum_{i=1}^n \left( \sum_{m=1}^M \mathbf{1}_{X_i \in Z_x^m}\right) } \nonumber \\
        & \hspace{-.5cm} \leq \frac{\sum_{i=1}^n L^2 \|X_i - x\|^{2\beta} \left( \sum_{m=1}^M \mathbf{1}_{X_i \in Z_x^m}\right) }{\sum_{i=1}^n \left( \sum_{m=1}^M \mathbf{1}_{X_i \in Z_x^m}\right) } \nonumber \\
        & \hspace{-.5cm} = L^2 \cdot \sum_{m=1}^M \left( \frac{\sum_{i=1}^n \mathbf{1}_{X_i \in Z_x^{m}}}{\sum_{m'=1}^M \left( \sum_{i=1}^n \mathbf{1}_{X_i \in Z_x^{m'}} \right)} \right) \operatorname{diam}(Z_x^{m} \cap \supp(\nu))^{2\beta}. \label{ln:bias_bounded_by_convex_combo}  
    \end{align}
    In the convex combination, the term $\operatorname{diam}(Z_x^{m} \cap \supp(\nu))^{2\beta}$ is weighted proportionally to the number of data points that fall in $Z_x^{m}$. The weights, therefore, are random variables depending on the training data and the partitions. Comparatively more weight is given to terms $\operatorname{diam}(Z_x^{m} \cap \supp(\nu))^{2\beta}$ for which $Z_x^{m}$ contains more data points, and we would expect to see more data points in cells with larger diameters if, for instance, $\nu$ were the uniform measure. Since a convex combination is no greater than the maximum quantity being combined,
    \begin{align}
        \sum_{m=1}^M \left( \frac{\sum_{i=1}^n \mathbf{1}_{X_i \in Z_x^{m}}}{\sum_{m'=1}^M \left( \sum_{i=1}^n \mathbf{1}_{X_i \in Z_x^{m'}} \right)} \right) \operatorname{diam}(Z_x^{m} \cap \supp(\nu))^{2\beta} \leq \max_{m=1}^M \operatorname{diam}(Z_x^{m} \cap \supp(\nu))^{2\beta}. \label{ln:bias_bounded_by_max_diameter}
    \end{align}
    
    We continue the analysis of the kernel estimator's risk, as decomposed in \cref{ln:bias_variance_decomposition}, by bounding its conditional variance. Using the bound on the noise in the training labels,
    \begin{align}
        & \E\left[ \left. \left( \hat{f}_{\mathcal{P}, M}(x) - \Bar{f}_{\mathcal{P}, M}(x) \right)^2 \mathbf{1}_{\exists i, m : X_i \in Z_x^{m}} \right \rvert \{X_i\}_{i=1}^n, \left\{\mathcal{P}^{(m)}\right\}_{m=1}^M \right] \nonumber \\
        & \hspace{.5cm} = \mathbf{1}_{\exists i, m : X_i \in Z_x^{m}} \cdot \operatorname{Var}\left[ \left. \hat{f}_{\mathcal{P}, M}(x) \right\rvert \{X_i\}_{i=1}^n, \left\{\mathcal{P}^{(m)}\right\}_{m=1}^M \right] \nonumber \\
        & \hspace{.5cm} = \mathbf{1}_{\exists i, m: X_i \in Z_x^{m}}  \frac{\sum_{i=1}^n \operatorname{Var}\left[\left. \xi_i \sum_{m=1}^M \mathbf{1}_{X_i \in Z_x^{m}} \right\rvert X_i, \left\{\mathcal{P}^{(m)}\right\}_{m=1}^M\right]}{\left( \sum_{i=1}^{n} \sum_{m=1}^M \mathbf{1}_{X_i \in Z_x^{m}}\right)^2} \nonumber \\
        & \hspace{.5cm} \leq \mathbf{1}_{\exists i, m: X_i \in Z_x^{m}} \cdot \sigma^2 \cdot \frac{\sum_{i=1}^n \left( \sum_{m=1}^M \mathbf{1}_{X_i \in Z_x^{m}} \right)^2}{\left( \sum_{i=1}^n \sum_{m=1}^M \mathbf{1}_{X_i \in Z_x^{m}} \right)^2} \nonumber \\
        & \hspace{.5cm} {\leq \mathbf{1}_{\exists i, m: X_i \in Z_x^{m}} \cdot \sigma^2 \cdot \frac{\sum_{i=1}^n M \sum_{m=1}^M \mathbf{1}_{X_i \in Z_x^{m}}}{\left( \sum_{i=1}^n \sum_{m=1}^M \mathbf{1}_{X_i \in Z_x^{m}} \right)^2}} \nonumber \\
        & \hspace{.5cm} {=  \mathbf{1}_{\exists i, m: X_i \in Z_x^{m}} \cdot \sigma^2 \cdot \frac{M}{\sum_{i=1}^n \sum_{m=1}^M \mathbf{1}_{X_i \in Z_x^{m}}}} \nonumber \\
        & \hspace{.5cm} {\leq \mathbf{1}_{\exists i, m: X_i \in Z_x^{m}} \cdot \sigma^2 \cdot \frac{M}{\sum_{i=1}^n \mathbf{1}_{X_i \in \bigcup_{m=1}^M Z_x^{m}}}} \label{ln:ensemble_conditional_variance}
    \end{align}
    The final two inequalities are based on the geometric observation 
    \[\mathbf{1}_{X_i \in \bigcup_{m=1}^M Z_x^{m}} \leq \sum_{m=1}^M \mathbf{1}_{X_i \in Z_x^{m}} \leq M.\]
    This step is useful in the analysis because it allows us to analyze the denominator as the sum of independent binomial random variables, conditioned on the partitions. Moreover, the variance is intuitively controlled by averaging many independent data points, and while this bound might be loose for large $M$, it preserves the overall number of independent data points used to form the local average $\hat{f}_{\mathcal{P}, M}(x)$. For notational brevity, let $Z_x^{[M]} := \bigcup_{m=1}^M Z_x^{m}$.
    
    Due to Hypothesis \ref{as:purely_random} and independence of $\{X_i\}_{i=1}^n$, the denominator in \cref{ln:ensemble_conditional_variance} is the sum of two independent binomial random variables, conditioned on the partitions. By \cref{lm:binomial_random_variables},
    \begin{align}
        & \E\left[ \left. \mathbf{1}_{\exists i, m: X_i \in Z_x^{m}} \frac{M \sigma^2}{\sum_{i=1}^{n_\nu} \mathbf{1}_{X_i \in Z_x^{[M]}} + \sum_{i = n_\nu + 1}^n \mathbf{1}_{X_i \in Z_x^{[M]}}} \right\rvert \left\{\mathcal{P}^{(m)}\right\}_{m=1}^M \right] \nonumber \\
        & \leq M \sigma^2 \cdot \E\left[ \left. 1 \wedge \frac{1}{\sum_{i=1}^{n_\nu} \mathbf{1}_{X_i \in Z_x^{[M]}} + \sum_{i = n_\nu + 1}^n \mathbf{1}_{X_i \in Z_x^{[M]}}} \right\rvert \left\{\mathcal{P}^{(m)}\right\}_{m=1}^M \right] \nonumber \\
        & \leq 4 M \sigma^2 \wedge \frac{18 M \sigma^2}{n_\nu \cdot \nu\left( Z_x^{[M]} \right) + n_\mu \cdot \mu\left( Z_x^{[M]} \right)}. \label{ln:ensemble_conditional_variance_bound}
    \end{align}
    
    So far, we have bounded the bias and variance in the case that for at least one partition, indexed $m$, there is at least one data point, indexed $i$, such that $X_i \in Z_x^{m}$, so that $\hat{f}_{\mathcal{P}, M}(x)$ is a local average of training labels. If not, then the values of $\hat{f}_{\mathcal{P}, M}(x)$ and $\Bar{f}_{\mathcal{P}, M}(x)$ are identically zero, so the squared error is $(f^*(x))^2$.
    Conditioned on the partitions, by independence of $\{X_i\}_i$, the probability of no data falling in the cells is
    \begin{align}
        (f^*(x))^2 \cdot \Prob\left[ \left. \nexists i, m: X_i \in Z_x^{m} \right\rvert \left\{\mathcal{P}^{(m)}\right\}_{m=1}^M \right] & = (f^*(x))^2 \left(1 - \nu\left( Z_x^{[M]} \right)\right)^{n_\nu} \left(1 - \mu\left( Z_x^{[M]} \right)\right)^{n_\mu} \nonumber \\
        & \leq (f^*(x))^2 \wedge \frac{(f^*(x))^2}{n_\nu \cdot \nu\left( Z_x^{[M]} \right) + n_\mu \cdot \mu\left( Z_x^{[M]} \right)}. \label{ln:error_when_no_local_data}
    \end{align}
    We used the identities $(1-p_1)^{n_1} (1-p_2)^{n_2} \leq \exp(-(n_1 p_1 + n_2 p_2)) \leq \frac{1}{n_1 p_1 + n_2 p_2}$, which hold for $p_1, p_2 \in (0, 1)$ and integers $n_1, n_2 \geq 0$. See footnote 3 in \cite{pathak_new_2022} for similar usage. We also bounded the probability above by $1$. We absorb \cref{ln:error_when_no_local_data} into the variance term.
    
    Finally, combining \cref{ln:bias_variance_decomposition}, \cref{ln:bias_bounded_by_convex_combo}, \cref{ln:bias_bounded_by_max_diameter}, \cref{ln:ensemble_conditional_variance_bound}, and \cref{ln:error_when_no_local_data}, 
    \begin{align}
        \E[(\hat{f}_{\mathcal{P}, M}(x) - f^*(x))^2] 
        & = \E\left[\left( \Bar{f}_{\mathcal{P}, M}(x) - f^*(x) \right)^2 \mathbf{1}_{\exists i, m : X_i \in Z_x^{m}}\right] \nonumber \\
        & \hspace{.5cm} + \E\left[ \E\left[ \left. \left( \hat{f}_{\mathcal{P}, M}(x) - \Bar{f}_{\mathcal{P}, M}(x) \right)^2 \mathbf{1}_{\exists i, m : X_i \in Z_x^{m}} \right \rvert \{X_i\}_i, \left\{\mathcal{P}^{(m)}\right\}_m \right] \right] \nonumber \\
        & \hspace{.5cm} + \E\left[ (f^*(x))^2 \cdot \Prob\left[ \nexists i, m: X_i \in Z_x^{m} \left\rvert \left\{\mathcal{P}^{(m)}\right\}_m \right. \right] \right] \nonumber \\
        & \hspace{-.5cm} \leq \E\left[ L^2 \max_{m=1}^M \operatorname{diam}(Z_x^{m} \cap \supp(\nu))^{2\beta} \right] \nonumber \\
        & + \E\left[ 4 M \sigma^2 \wedge \frac{18 M \sigma^2}{n_\nu \cdot \nu\left( Z_x^{[M]} \right) + n_\mu \cdot \mu\left( Z_x^{[M]} \right)} \right] \nonumber \\
        & + \E\left[ (f^*(x))^2 \wedge \frac{(f^*(x))^2}{n_\nu \cdot \nu\left( Z_x^{[M]} \right) + n_\mu \cdot \mu\left( Z_x^{[M]} \right)} \right] \nonumber \\
        & \hspace{-.5cm} \leq L^2 \cdot \E\left[ \max_{m=1}^M \operatorname{diam}(Z_x^{m} \cap \supp(\nu))^{2\beta} \right] \nonumber \\
        & + \frac{18 M \sigma^2 + (f^*(x))^2}{n} \cdot \E\left[n \wedge \frac{1}{\frac{n_\nu}{n} \cdot \nu\left( Z_x^{[M]} \right) + \frac{n_\mu}{n} \cdot \mu\left( Z_x^{[M]} \right)} \right]. \nonumber
    \end{align}
\end{proof}

\section{Proof of \Cref{lm:STIT_bias_bound}} \label{sc:STIT_bias_bound_proof}

\begin{proof}[Proof of \cref{lm:STIT_bias_bound}]
    The STIT tessellation has stationarity and the scaling property, so $Z_x^{m, \lambda}$ is distributed as $\frac{1}{\lambda} (Z_0^{m, 1}) + x$. Translation does not change a convex body's diameter. We use these observations to separate the bias into factors depending on $\lambda$ and $M$:
    \begin{align}
        \E\left[ \max_{m=1}^M \operatorname{diam}(Z_x^{m, \lambda})^{2\beta} \right] 
        & = \E\left[ \max_{m=1}^M \operatorname{diam}\left(\frac{Z_0^{m, 1}}{\lambda}\right)^{2\beta} \right] = \frac{1}{\lambda^{2\beta}} \cdot \E\left[ \max_{m=1}^M \operatorname{diam}(Z_0^{m, 1})^{2\beta} \right]. \label{ln:bias_dependence_on_lambda}
    \end{align}
    
    To clarify the dependence of $\E\left[ \max_{m=1}^M \operatorname{diam}(Z_0^{m, 1})^{2\beta} \right]$ on $M$, we obtain a moment bound for $\max_{m=1}^M \operatorname{diam}(Z_0^{m, 1})$ via a previously obtained tail bound on $\operatorname{diam}(Z_0^{m, 1})$. %
    
    We note that we can apply known results about the distribution of the zero cell of the Poisson hyperplane tessellation because the zero cell of the STIT tessellation has the same distribution as that of the Poisson hyperplane tessellation with corresponding parameters, as argued in \cite{oreilly_minimax_2024} by citing Corollary 1 of \cite{Schreiber_Thäle_2013} and Theorem 10.4.1 of \cite{Schneider_Weil_2008}. We work from the proof of Theorem 6.1.2 in \cite{hug_poisson_2024}, which is provided in the text after the statement of Theorem 6.3.1, a more general result. The authors identify a constant $c_\Pi > 0$, which depends on the zonoid $\Pi$, such that, for any $m$,
    \begin{equation}
        \Prob\left[ \frac{1}{2} \operatorname{diam}(Z_0^{m, 1}) \geq t \right] \leq 2d \cdot \exp(-c_\Pi t). \label{ln:zero_cell_diameter_tail_bound}
    \end{equation}
    More precisely, Theorem 6.1.2 provides a right-sided tail bound for the radius of a sphere containing $Z_0^{m, 1}$, hence our inclusion of the $\frac{1}{2}$ in the line above.

    For $dM \geq 2$, we follow the proof technique of Proposition 2.7.6 in \cite{Vershynin_2026}, to derive a tail bound for $\max_{m=1}^M \operatorname{diam}(Z_0^{m, 1})$. Taking a union bound, applying \cref{ln:zero_cell_diameter_tail_bound}, and appealing to a numerical argument given in \Cref{sc:moment_bound_from_tail_bound}, we find
    \begin{equation}
        \Prob\left[ \max_{m=1}^M \operatorname{diam}(Z_0^{m, 1}) \geq t \right] 
        \leq 2 \exp\left(-\frac{c_\Pi t}{6 \log (dM)}\right). \label{ln:max_zero_cell_diameter_tail_bound}
    \end{equation}
    From \cref{ln:max_zero_cell_diameter_tail_bound} (and \cref{ln:zero_cell_diameter_tail_bound} in the case $d = M = 1$), we obtain the moment bound
    \begin{equation} \label{ln:max_zero_cell_moment_bound_general}
        \E\left[ \max_{m=1}^M \operatorname{diam}(Z_0^{m, 1})^{2\beta} \right] \leq \left(\E\left[ \left( \max_{m=1}^M \operatorname{diam}\left(Z_0^{m, 1}\right)\right)^{2} \right]\right)^{\beta} \lesssim \frac{1}{c_\Pi^{2\beta}} \cdot \left(1 \vee \log^{2\beta} (dM)\right).
    \end{equation}
    The $\lesssim$ elides a universal constant. Details for the derivation of \cref{ln:max_zero_cell_moment_bound_general} from \cref{ln:max_zero_cell_diameter_tail_bound} and \cref{ln:zero_cell_diameter_tail_bound} are provided in \Cref{sc:moment_bound_from_tail_bound}. The result for a general STIT tessellation is obtained by substituting \cref{ln:max_zero_cell_moment_bound_general} into \cref{ln:bias_dependence_on_lambda}. For the statement of the lemma, we refer to $c_\Pi^{-2\beta}$ as $c_{\beta, \Pi}$.

    The proof for the Mondrian case is provided in \Cref{sc:proof_Mondrian_bias_bound}.
\end{proof}

\section{Proof of \Cref{pr:sufficient_ensemble_size}} \label{sc:sufficient_ensemble_size_proof}

\begin{proof}[Proof of \cref{pr:sufficient_ensemble_size}]
    Per \cref{cr:STIT_convergence_rate}, it suffices to prove that $\E\left[ n \wedge \frac{1}{\nu\left( \bigcup_{m=1}^M Z_x^{m, \lambda} \right)} \right]$ is on the order of $\lambda^\alpha$.
    
    As in \cref{ln:stit_zero_cell_union} and \cref{ln:mondrian_zero_cell_union}, we begin by rewriting the integrated $\mathcal{P}^M_\lambda$-transfer discrepancy in terms unit exponential random variables $\{T^+_m, T^-_m\}_{m=1}^M$. From the distribution of the cell of the Mondrian \cite{Mourtada_Gaiffas_Scornet_2020},
    \begin{equation} \label{ln:rewrite_cell_with_exponentials}
        \E\left[n \wedge \frac{1}{\nu\left(\bigcup_{m=1}^M Z_X^{m, \lambda}\right)}\right] = \E\left[n \wedge \frac{1}{\nu\left(\left[-\frac{1}{\lambda}\max_{m=1}^M T^-_m + X, \frac{1}{\lambda}\max_{m=1}^M T^+_m + X\right]\right)}\right].
    \end{equation}
    
    Inspired by the proof technique in Appendix C of \cite{Galbraith_Kpotufe_2024}, we condition on $\{T^+_m, T^-_m\}_{m=1}^M$ and bifurcate the support of $X$ into a region closer to the $\nu$ singularity at $0$ and a region further away. To account for the boundaries of the support, we will compute the measure $\nu$ of only a subset of each cell. Unlike their proof, however, the extents of the cell on either side of $X$ are random and independent, so there is a nonzero probability that $X$ falls into neither region. We will bound this probability and show that, even when multiplied by the truncation value $n$, suitable values of $M$ and $\lambda$ can ensure this term quickly approaches zero. For compactness, notate $T^-_{[M]} := \max_{m=1}^M T^-_m$ and $T^+_{[M]} := \max_{m=1}^M T^+_m$. Continuing to analyze \cref{ln:rewrite_cell_with_exponentials},
    \begin{align}
    	& \E \left[n \wedge \frac{1}{\nu\left(\left[-\frac{1}{\lambda}T^-_{[M]} + X, \frac{1}{\lambda}T^+_{[M]} + X\right]\right)}\right] \nonumber \\
    	& \leq \E_{\{T^+_m, T^-_m\}_m} \left[ \E_X \left[\frac{\mathbf{1}_{X \leq \left(\frac{2}{\lambda} T^-_{[M]}\right) \wedge 1} \cdot \mathbf{1}_{X \leq 1 - \frac{1}{\lambda}T^+_{[M]}}}{\nu\left(\left[-\frac{1}{\lambda}T^-_{[M]} + X, \frac{1}{\lambda}T^+_{[M]} + X\right]\right)} + \frac{\mathbf{1}_{X > \left(\frac{2}{\lambda} T^-_{[M]}\right) \wedge 1}}{\nu\left(\left[-\frac{1}{\lambda}T^-_{[M]} + X, \frac{1}{\lambda}T^+_{[M]} + X\right]\right)}\right] \right] \nonumber \\
        & \hspace{.5cm} + n \cdot \Prob\left[ 1 - \frac{1}{\lambda}T^+_{[M]} < X \leq \left(\frac{2}{\lambda} T^-_{[M]}\right) \wedge 1 \right] \nonumber \\
    	& \leq \E_{\{T^+_m, T^-_m\}_m} \left[ \E_X \left[\frac{\mathbf{1}_{X \leq \left(\frac{2}{\lambda} T^-_{[M]}\right) \wedge \left(1 - \frac{1}{\lambda} T^+_{[M]} \right)}}{\nu\left(\left[X, \frac{1}{\lambda}T^+_{[M]} + X\right]\right)}\right] \right] + \E_{\{T^+_m, T^-_m\}_m} \left[ \E_X \left[\frac{\mathbf{1}_{X > \left(\frac{2}{\lambda} T^-_{[M]}\right) \wedge 1}}{\nu\left(\left[-\frac{1}{\lambda}T^-_{[M]} + X, X\right]\right)}\right] \right] \nonumber \\
        & \hspace{.5cm} + n \cdot \Prob\left[ 1 - \frac{1}{\lambda}T^+_{[M]} < \left(\frac{2}{\lambda} T^-_{[M]}\right) \wedge 1 \right]. \label{ln:bifurcated_interval}
    \end{align}
    
    The second term in \cref{ln:bifurcated_interval} encompasses values of $X$ that are far enough from $0$ that we can lower-bound the density of $\nu$ on the interval $\left[-\frac{1}{\lambda}T^-_{[M]} + X, X\right]$, which is a subset of the cell $\left[-\frac{1}{\lambda}T^-_{[M]} + X, \frac{1}{\lambda}T^+_{[M]} + X\right]$ and of $\supp(\nu) = [0, 1]$. For $x > \frac{2}{\lambda} T^-_{[M]}$, any $x' \in \left[-\frac{1}{\lambda}T^-_{[M]} + x, x\right]$ satisfies
    \[ x' \geq x - \frac{1}{\lambda}T^-_{[M]} = x \left( 1 - \frac{1}{x} \cdot \frac{1}{\lambda}T^-_{[M]} \right) \geq x \left(1 - \frac{1}{2}\right) = \frac{x}{2}, \]
    where the last inequality holds because $x > \frac{2}{\lambda} T^-_{[M]}$ can be rearranged to $\frac{1}{2} > \frac{1}{x} \cdot \frac{1}{\lambda}T^-_{[M]}$. Thus, the density of $\nu$ is at least $(\alpha+1) \left(\frac{x}{2}\right)^\alpha$ at any such $x'$. In summary, with $X$ in place of $x$,
    \[ X > \frac{2}{\lambda} T^-_{[M]} \implies \nu\left( \left[-\frac{1}{\lambda}T^-_{[M]} + X, X\right] \right) \geq \frac{1}{\alpha+1} \left(\frac{X}{2}\right)^\alpha \cdot \frac{1}{\lambda}T^-_{[M]} \]
    We next integrate with respect to $X$, deriving an expression for the inner expectation of the second term in \cref{ln:bifurcated_interval}. Using the assumption that $\alpha - 1$ is strictly positive,
    \begin{align}
    	\E_X \left[\frac{\mathbf{1}_{X > \left(\frac{2}{\lambda} T^-_{[M]}\right) \wedge 1}}{\nu\left(\left[-\frac{1}{\lambda}T^-_{[M]} + X, X\right]\right)}\right] & = \int_{\left(\frac{2}{\lambda} T^-_{[M]}\right) \wedge 1}^1 \frac{dx}{\nu\left(\left[-\frac{1}{\lambda}T^-_{[M]} + x, x\right]\right)} \nonumber \\
    	& \leq \int_{\left(\frac{2}{\lambda} T^-_{[M]}\right) \wedge 1}^1 \frac{dx}{(\alpha+1) \left(\frac{x}{2}\right)^\alpha \cdot \frac{1}{\lambda}T^-_{[M]}} \nonumber \\
    	& = 2^\alpha \cdot \frac{1}{(\alpha - 1)(\alpha + 1)} \cdot \frac{\lambda}{T^-_{[M]}} \cdot \left( \frac{1}{\left( \left(\frac{2}{\lambda} T^-_{[M]}\right) \wedge 1 \right)^{\alpha - 1}} - 1 \right) \nonumber \\
    	& \leq 2 \cdot \frac{1}{(\alpha - 1)(\alpha + 1)} \cdot \frac{\lambda^\alpha}{\left( T^-_{[M]} \right)^\alpha}. \label{ln:upper_region_x_integral}
    \end{align}
    The last inequality above holds because if $\frac{2}{\lambda} T^-_{[M]} > 1$, then the expression on the second-to-last line is $0$ and, otherwise, $\left(\frac{2}{\lambda} T^-_{[M]}\right) \wedge 1$ takes the value $\frac{2}{\lambda} T^-_{[M]}$. In preparation to integrate with respect to $\{T^-_m\}_m$, we lower-bound the random variable $\left( T^-_{[M]} \right)^\alpha$ by a scaling of $\left( \sum_{m=1}^M T^-_m \right)^\alpha$. By definition of $T^-_{[M]}$,
    \begin{equation}
        \left( \sum_{m=1}^M T^-_m \right)^\alpha 
        \leq \left( \sum_{m=1}^M T^-_{[M]} \right)^\alpha = M^\alpha \left(T^-_{[M]}\right)^\alpha \label{ln:lower_bound_max_with_sum}
    \end{equation}
    $\sum_{m=1}^M T^-_m$ has distribution $\operatorname{Gamma}(M, 1)$, whose density is known. We compute its moment of order $(-\alpha)$. Since $M \geq \alpha + 1$,
    \begin{equation}
        \E_{\{T^-_m\}_m} \left[ \frac{1}{\left( \sum_{m=1}^M T^-_m \right)^\alpha} \right] = \int_0^\infty \frac{1}{t^\alpha} \cdot \frac{t^{M-1} e^{-t}}{\Gamma(M)} \; dt = \frac{1}{\Gamma(M)} \int_0^\infty t^{M - \alpha - 1} e^{-t} \; dt = \frac{(M - \alpha - 1)!}{(M - 1)!}. \label{ln:finite_gamma_inverse_moment}
    \end{equation}
    Combining \cref{ln:upper_region_x_integral}, \cref{ln:lower_bound_max_with_sum}, and \cref{ln:finite_gamma_inverse_moment}, and noting that $\alpha \geq 2$ by assumption, 
    \begin{align}
        \E_{\{T^-_m\}_m} \left[ \E_X \left[\frac{\mathbf{1}_{X > \left(\frac{2}{\lambda} T^-_{[M]}\right) \wedge 1}}{\nu\left(\left[-\frac{1}{\lambda}T^-_{[M]} + X, X\right]\right)}\right] \right] & \leq \E_{\{T^-_m\}_m} \left[ 2 \cdot \frac{1}{(\alpha - 1)(\alpha + 1)} \cdot \frac{\lambda^\alpha}{\left( T^-_{[M]} \right)^\alpha}  \right] \nonumber \\
        & \hspace{-1cm} \leq 2 \cdot \frac{1}{(\alpha - 1)(\alpha + 1)} \cdot \lambda^\alpha \cdot M^{\alpha} \cdot \E_{\{T^-_m\}_m} \left[ \frac{1}{\left( \sum_{m=1}^M T^-_m \right)^\alpha} \right] \nonumber \\
        & \hspace{-1cm} = 2 \cdot \frac{1}{(\alpha - 1)(\alpha + 1)} \cdot \lambda^\alpha \cdot M^{\alpha} \cdot \frac{(M - \alpha - 1)!}{(M - 1)!} \nonumber \\
        & \hspace{-1cm} \leq \frac{\lambda^\alpha}{\left( 1 - \frac{\alpha}{M} \right)^\alpha}. \label{ln:upper_region_upper_bound}
    \end{align}
    This completes an expression for the second term of \cref{ln:bifurcated_interval}.
    
    We turn now to the first term in \cref{ln:bifurcated_interval}. It concerns the region of the support of $X$ that is so close to the singularity of $\nu$ at $0$ that we cannot lower-bound the density of $\nu$. Rather, we will use the rate given by the $\mathcal{P}^M_\lambda$-transfer exponent, which is $\lambda^{\alpha + 1}$, and show that this region is only on the order of $\frac{1}{\lambda}$ in width, recovering the rate $\lambda^\alpha$ as desired. To begin, because the density of $\nu$ is increasing on its support,
    \[\nu\left(\left[X, \frac{1}{\lambda}T^+_{[M]} + X\right]\right) = \nu\left(\left[X, \left(\frac{1}{\lambda}T^+_{[M]} + X\right) \wedge 1\right]\right) \geq \nu\left(\left[0, \left(\frac{1}{\lambda}T^+_{[M]}\right) \wedge (1 - X) \right] \right).\]
    The numerator of the first term in \cref{ln:bifurcated_interval} indicates that $X \leq \left(\frac{2}{\lambda} T^-_{[M]}\right) \wedge \left(1 - \frac{1}{\lambda} T^+_{[M]} \right)$. In particular, we have $\left(\frac{1}{\lambda}T^+_{[M]}\right) \leq 1 - X$ and so,
    \[\nu\left(\left[0, \left(\frac{1}{\lambda}T^+_{[M]}\right) \wedge (1 - X) \right] \right) = \nu\left(\left[0, \frac{1}{\lambda}T^+_{[M]} \right] \right) = \left( \frac{1}{\lambda}T^+_{[M]} \right)^{\alpha+1} \wedge 1.\]
    We integrate with respect to $X$ to derive an expression for the inner expectation of the first term of \cref{ln:bifurcated_interval}. Using independence of $\{T^+_m\}_m$ and $\{T^-_m\}_m$ (and dropping indicator functions after they have served their purpose),
    \begin{align}
        \E_X \left[\frac{\mathbf{1}_{X \leq \left(\frac{2}{\lambda} T^-_{[M]}\right) \wedge \left(1 - \frac{1}{\lambda} T^+_{[M]} \right)}}{\nu\left(\left[X, \frac{1}{\lambda}T^+_{[M]} + X\right]\right)}\right] & \leq \E_X \left[\frac{\mathbf{1}_{X \leq \frac{2}{\lambda} T^-_{[M]}}}{\left( \frac{1}{\lambda}T^+_{[M]} \right)^{\alpha+1} \wedge 1}\right]  \leq \frac{\frac{2}{\lambda} T^-_{[M]}}{\left( \frac{1}{\lambda}T^+_{[M]} \right)^{\alpha+1} \wedge 1}. \label{ln:lower_region_conditional_upper_bound}
    \end{align}
    The numerator and denominator in \cref{ln:lower_region_conditional_upper_bound} are independent. A union bound to controls the expectation of the numerator by 
    \begin{equation}
        \E\left[ \frac{2}{\lambda} T^-_{[M]} \right] = \frac{2}{\lambda} \int_0^\infty \Prob\left[T^-_{[M]} \geq t\right] \; dt \leq \frac{2}{\lambda} \int_0^\infty \sum_{m=1}^M \Prob\left[T^-_m \geq t\right] \; dt = \frac{2M}{\lambda}. \label{ln:lower_region_numerator_upper_bound}
    \end{equation}
    To compute the expectation of (the reciprocal of) the denominator in \cref{ln:lower_region_conditional_upper_bound}, we break into the cases $T^+_{[M]} \geq \lambda$ and $T^+_{[M]} < \lambda$. For the latter case, noting that $M \geq \alpha + 2$ and that $T^+_m$ and $T^-_m$ are identically distributed, we apply \cref{ln:lower_bound_max_with_sum} and \cref{ln:finite_gamma_inverse_moment} with $\alpha+1$ in place of $\alpha$ to obtain
    \begin{align*}
        \E\left[ \frac{1}{\left( \frac{1}{\lambda}T^+_{[M]} \right)^{\alpha+1} \wedge 1} \right] & = \E\left[ \frac{\mathbf{1}_{T^+_{[M]} \geq \lambda}}{\left( \frac{1}{\lambda}T^+_{[M]} \right)^{\alpha+1} \wedge 1} \right] + \E\left[ \frac{\mathbf{1}_{T^+_{[M]} < \lambda}}{\left( \frac{1}{\lambda}T^+_{[M]} \right)^{\alpha+1} \wedge 1} \right] \\
        & \leq \E\left[ \frac{\mathbf{1}_{T^+_{[M]} \geq \lambda}}{1} \right] + \E\left[ \frac{1}{\left( \frac{1}{\lambda}T^+_{[M]} \right)^{\alpha+1}} \right] \\
        & \leq \sum_{m=1}^M \Prob\left[T^-_m \geq \lambda\right] + \lambda^{\alpha + 1} \cdot \E\left[\frac{M^{\alpha+1}}{\left(\sum_{m=1}^M T^+_m\right)^{\alpha+1}}\right] \\
        & \leq Me^{-\lambda} + \lambda^{\alpha + 1} \cdot M^{\alpha+1} \cdot \frac{(M - \alpha - 2)!}{(M - 1)!} \\
        & \leq Me^{-\lambda} + \frac{\lambda^{\alpha+1}}{\left( 1 - \frac{\alpha+1}{M} \right)^{\alpha+1}}.
    \end{align*}
    This, together with \cref{ln:lower_region_conditional_upper_bound} and \cref{ln:lower_region_numerator_upper_bound}, gives us a bound on the first term of \cref{ln:bifurcated_interval}:
    \begin{align}
        \E_{\{T^+_m, T^-_m\}_m} \left[ \E_X \left[\frac{\mathbf{1}_{X \leq \left(\frac{2}{\lambda} T^-_{[M]}\right) \wedge \left(1 - \frac{1}{\lambda} T^+_{[M]} \right)}}{\nu\left(\left[X, \frac{1}{\lambda}T^+_{[M]} + X\right]\right)}\right] \right] & \leq \E_{\{T^+_m, T^-_m\}_m} \left[ \frac{\frac{2}{\lambda} T^-_{[M]}}{\left( \frac{1}{\lambda}T^+_{[M]} \right)^{\alpha+1} \wedge 1} \right] \nonumber \\
        & \leq \E\left[ \frac{2}{\lambda} T^-_{[M]} \right] \cdot \E\left[ \frac{1}{\left( \frac{1}{\lambda}T^+_{[M]} \right)^{\alpha+1} \wedge 1} \right] \nonumber \\
        & \leq \frac{2M}{\lambda} \left(Me^{-\lambda} + \frac{\lambda^{\alpha+1}}{\left( 1 - \frac{\alpha+1}{M} \right)^{\alpha+1}}\right) \nonumber \\
        & = 2M^2 \cdot \frac{e^{-\lambda}}{\lambda} + \lambda^{\alpha} \cdot \frac{2M}{\left( 1 - \frac{\alpha+1}{M} \right)^{\alpha+1}}. \label{ln:lower_region_upper_bound}
    \end{align}

    Finally, there is a chance that $X$ falls in neither of these regions but rather in the interval between, namely $\left( 1 - \frac{1}{\lambda} T^+_{[M]}, \left(\frac{2}{\lambda} T^-_{[M]}\right) \wedge 1 \right]$. We show that with high probability this interval is empty, meaning the lower endpoint exceeds the upper. By independence of $T^-_{[M]}$ and $T^+_{[M]}$, 
    \begin{align}
        \Prob\left[ 1 - \frac{1}{\lambda}T^+_{[M]} < \frac{2}{\lambda} T^-_{[M]} \right] & = \E_{\{T^+_m\}_m} \E_{\{T^-_m\}_m} \left[ \mathbf{1}_{T^-_{[M]} > \frac{1}{2}\left(\lambda - T^+_{[M]}\right)} \right] \nonumber \\
        & \leq \E_{\{T^+_m\}_m} \left[ \left( M \cdot \exp\left(-\frac{1}{2}\left(\lambda - T^+_{[M]}\right) \right) \right) \wedge 1 \right] \nonumber \\
        & \leq M \cdot \exp\left(-\frac{\lambda}{2}\right) \cdot \E\left[ \exp\left(\frac{1}{2} T^+_{[M]} \right) \right] \nonumber \\
        & \leq 2M^2 \cdot \exp\left(-\frac{\lambda}{2}\right). \label{ln:region_between_probability_bound}
    \end{align}
    The last inequality comes from manipulating the moment generating function of the unit exponential distribution, which is $t \mapsto \frac{1}{1-t}$ for $|t| < 1$, as follows:
    \[\E\left[ \exp\left(\frac{T^+_{[M]}}{2} \right) \right] = \E\left[\max_{m=1}^M \exp\left(\frac{T^+_m }{2} \right) \right] \leq \E\left[\sum_{m=1}^M \exp\left(\frac{T^+_m}{2}  \right) \right] = \sum_{m=1}^M \E\left[\exp\left(\frac{T^+_m}{2}  \right) \right] = 2M.\]

    Altogether, \cref{ln:bifurcated_interval}, \cref{ln:upper_region_upper_bound}, \cref{ln:lower_region_upper_bound}, and \cref{ln:region_between_probability_bound} imply 
    \[\E\left[n \wedge \frac{1}{\nu\left(\bigcup_{m=1}^M Z_X^{m, \lambda}\right)}\right] \leq \frac{2M^2 e^{-\lambda}}{\lambda} + \frac{\lambda^{\alpha} \cdot 2M}{\left( 1 - \frac{\alpha+1}{M} \right)^{\alpha+1}} + \frac{\lambda^\alpha}{\left( 1 - \frac{\alpha}{M} \right)^\alpha} + 2nM^2 e^{-\frac{\lambda}{2}}. \]
    
    Applying \cref{tm:integrated_STIT_risk_bound} and the Mondrian case from \cref{lm:STIT_bias_bound} gives the result. 
\end{proof}

\section{Lemma about Binomial Random Variables} 

To prove \cref{tm:pointwise_risk_bound}, we need the following technical lemma about binomial random variables. It is a special case of Lemma SA3 in \cite{Cattaneo_Klusowski_Underwood_2025} with slightly better constants. Lemma 6 in \cite{pathak_new_2022} provides a similar result, but it omits the truncation factor, which we need in \cref{pr:sufficient_ensemble_size}, and we found its proof somewhat unclear.

\begin{lemma} \label{lm:binomial_random_variables}
    Let $N_1$ and $N_2$ be independent binomial random variables with distributions $\operatorname{Binomial}(n_1, p_1)$ and $\operatorname{Binomial}(n_2, p_2)$, respectively, for positive integers $n_1, n_2$ and $p_1, p_2 \in [0, 1]$. Then,
    \[\E\left[1 \wedge \frac{1}{N_1 + N_2}\right] \leq 4 \wedge \frac{18}{n_1 p_1 + n_2 p_2}\]
\end{lemma}

\begin{proof}
    As in Lemma SA3 in \cite{Cattaneo_Klusowski_Underwood_2025}, we apply Bernstein's inequality (the one-tailed version of Proposition 2.9.5 in \cite{Vershynin_2026}) to a binomial random variable $N \sim \operatorname{Binomial}(n, p)$, negated. $-N$ is the sum of negated independent Bernoulli trials, denoted $\{U_1, ..., U_n\}$, so Bernstein's inequality applies to the centered sum $np - N = \sum_{i=1}^n (p - U_i)$. Each $p - U_i$ has mean 0, second moment no greater than $p$, and maximum absolute value no greater than $1$. Hence, Bernstein's inequality states that for all positive $t$,
    \[\Prob\left[np - N \geq t\right] \leq \exp\left(-\frac{\frac{1}{2}t^2}{np + \frac{1}{3}t}\right)\]
    Setting $t$ to $\frac{3np}{4}$, we obtain the following bound on the left tail of $N$.
    \[\Prob\left[N \leq \frac{np}{4}\right] \leq \exp\left(-\frac{\frac{1}{2}\left(\frac{3np}{4}\right)^2}{np + \frac{1}{3}\left(\frac{3np}{4}\right)}\right) = \exp\left(-\frac{9}{40}np\right)\]
    Then, we bifurcate into the cases $N \leq \frac{np}{4}$ and $N > \frac{np}{4}$ to analyze the truncated and shifted inverse moment. For any $a \geq 0$,
    \begin{align*}
        \E\left[ \frac{1}{1 \vee (N + a)} \right] & \leq \exp\left(-\frac{9}{40}np\right) \cdot \frac{1}{1 \vee a} + \frac{1}{1 \vee \left(\frac{np}{4} + a\right)} \\
        & \leq \frac{1}{1 \vee \left( \frac{9}{40}np + a \right)} + \frac{1}{1 \vee \left(\frac{np}{4} + a\right)} \\
        & \leq 2 \cdot \frac{1}{1 \vee \left( \frac{9}{40}np + a \right)}
    \end{align*}
    The second inequality above uses the numeric inequality $e^{-t} \cdot \frac{1}{1 \vee a} \leq 1 \wedge \frac{1}{t+a}$ for $a, t \geq 0$, which follows from the identity $e^t \leq \frac{1}{1+t}$ through the following argument. If $a \leq 1$, then $e^{-t} \cdot \frac{1}{1 \vee a} = e^{-t} \leq \frac{1}{t+1} \leq \frac{1}{t+a}$; also, $e^{-t} \cdot \frac{1}{1 \vee a} = e^{-t} \leq 1$. On the other hand, if $a > 1$, then $e^{-t} \leq \frac{1}{1+t} \leq \frac{1}{1 + \frac{t}{a}}$, so $e^{-t} \cdot \frac{1}{1 \vee a} = e^{-t} \cdot \frac{1}{a} \leq \frac{1}{1 + \frac{t}{a}} \cdot \frac{1}{a} = \frac{1}{t + a}$; also $e^{-t} \cdot \frac{1}{1 \vee a} = e^{-t} \cdot \frac{1}{a} \leq 1$.
    
    We use this tail bound for $N_1$ and $N_2$ in sequence, using independence to evaluate the expectation with respect to $N_1$ while conditioning on $N_2$.
    \begin{align*}
        \E\left[1 \wedge \frac{1}{N_1 + N_2}\right] & = \E_{N_2} \left[ \E_{N_1}\left[1 \wedge \frac{1}{N_1 + N_2}\right] \right] \\
        & \leq \E_{N_2} \left[ 2 \cdot \frac{1}{1 \vee \left( \frac{9}{40} n_1 p_1 + N_2 \right)} \right] \\
        & \leq 4 \cdot \frac{1}{1 \vee \left( \frac{9}{40} n_1 p_1 + \frac{9}{40} n_2 p_2 \right)} \\
        & = 4 \wedge \frac{18}{n_1 p_1 + n_2 p_2}
    \end{align*}
\end{proof}

\section{Proof of Mondrian Kernel Estimator Bias Bound} \label{sc:proof_Mondrian_bias_bound}

\begin{proof}[Proof of Mondrian case in \cref{lm:STIT_bias_bound}]
    The Mondrian case can be handled differently to obtain an explicit dependence on $d$ and $\beta$.

    Since $t \mapsto t^{\beta}$ is concave, we apply Jensen's inequality as before to isolate the second moment, which we will subsequently bound.
    \[\E\left[ L^2 \max_{m=1}^M \operatorname{diam}(Z_x^{m, \lambda} \cap \supp(\nu))^{2\beta} \right] \leq L^2 \left(\E\left[ \max_{m=1}^M \operatorname{diam}(Z_x^{m, \lambda})^{2} \right]\right)^{\beta}\]
    For a given $m$, the squared diameter of $Z_x^{m, \lambda}$ is stochastically upper-bounded by the sum of $d$ squared, independent $\operatorname{Gamma}(2, \lambda)$-distributed variables, per the proof of Corollary 1 in \cite{Mourtada_Gaiffas_Scornet_2020}. Letting $\{W_{j, m}\}$ and $\{W_{m}\}$ refer to independent $\operatorname{Gamma}(2, 1)$ variables, we can bound the second moment of $\operatorname{diam}(Z_x^{m, \lambda})$.
    \[\E\left[ \max_{m=1}^M \operatorname{diam}(Z_x^{m, \lambda})^{2} \right] \leq \E\left[ \max_{m=1}^M \sum_{j=1}^d \frac{W_{j, m}^2}{\lambda^2} \right] \leq \E\left[ \sum_{j=1}^d \max_{m=1}^M \frac{W_{j, m}^2}{\lambda^2} \right] = \frac{d}{\lambda^2} \cdot \E\left[ \max_{m=1}^M W_{m}^2 \right] = \frac{d}{\lambda^2} \cdot \E\left[ \left(\max_{m=1}^M W_{m}\right)^2 \right]\]

    Each $W_m$ is Gamma-distributed, hence sub-exponential \cite{Vershynin_2026}. We adapt the (second) proof of Proposition 2.7.6 in \cite{Vershynin_2026} to show that the maximum of $M$ sub-exponential random variables is sub-exponential with sub-exponential norm on the order of $\log M$. From this we infer a second-moment bound for $\max_{m=1}^M W_{m}$, as desired.

    For each $m$, the variable $W_m$ has sub-exponential norm $\frac{2}{2-\sqrt{2}}$ because the moment generating function of $W_m$, evaluated at the reciprocal of this number, is $2$. (See the definition in line (2.25) in \cite{Vershynin_2026}.)
    \[\E\left[\exp\left(\frac{W_m}{\frac{2}{2 - \sqrt{2}}}\right)\right] = \left(1 - \frac{2-\sqrt{2}}{2}\right)^{-2} = 2\]
    From this, we derive a similar bound on the moment generating function of $\max_{m=1}^M W_{m}$, after scaling by a logarithmic factor.
    \begin{align*}
        \E\left[ \exp\left( \frac{\max_{m=1}^M W_{m}}{\frac{2}{2 - \sqrt{2}} \cdot 2 \log (2M)} \right) \right] & \leq \left( \E\left[ \exp\left( \max_{m=1}^M \frac{W_{m}}{\frac{2}{2 - \sqrt{2}}} \right) \right] \right)^{\frac{1}{2 \log (2M)}} \\
        & = \left( \E\left[ \max_{m=1}^M \exp\left( \frac{W_{m}}{\frac{2}{2 - \sqrt{2}}} \right) \right] \right)^{\frac{1}{2 \log (2M)}} \\
        & \leq \left( \E\left[ \sum_{m=1}^M \exp\left( \frac{W_{m}}{\frac{2}{2 - \sqrt{2}}} \right) \right] \right)^{\frac{1}{2 \log (2M)}} \\
        & \leq \left( M \cdot 2 \right)^{\frac{1}{2 \log (2M)}} = \sqrt{e} \leq 2
    \end{align*}
    The first inequality above is Jensen's inequality for concave functions; it holds for $M \geq 2$, where $t \mapsto t^{\frac{1}{2 \log (2M)}}$ is concave.

    Adapting the proof of Proposition 2.6.1 in \cite{Vershynin_2026}, we obtain a sub-exponential tail bound for the random variable $\frac{\max_{m=1}^M W_{m}}{\frac{2}{2 - \sqrt{2}} \cdot 2 \log (2M)}$.
    \[\Prob\left[ \frac{\max_{m=1}^M W_{m}}{\frac{2}{2 - \sqrt{2}} \cdot 2 \log (2M)} > t \right] = \Prob\left[ \exp\left( \frac{\max_{m=1}^M W_{m}}{\frac{2}{2 - \sqrt{2}} \cdot 2 \log (2M)} \right) > e^t \right] \leq e^{-t} \cdot \E\left[ \exp\left( \frac{\max_{m=1}^M W_{m}}{\frac{2}{2 - \sqrt{2}} \cdot 2 \log (2M)} \right) \right] \leq 2 e^{-t} \]
    Finally, we derive a second moment bound for $\max_{m=1}^M W_{m}$.
    \begin{align*}
        \E\left[ \left( \frac{\max_{m=1}^M W_{m}}{\frac{2}{2 - \sqrt{2}} \cdot 2 \log (2M)} \right)^2 \right] & = \int_0^\infty \Prob\left[ \frac{\max_{m=1}^M W_{m}}{\frac{2}{2 - \sqrt{2}} \cdot 2 \log (2M)} > t \right] \cdot 2t \; dt \leq 4 \int_0^\infty t e^{-t} \; dt = 4 \\
        \E\left[ \left( \max_{m=1}^M W_{m} \right)^2 \right] & = 4 \left(\frac{2 - \sqrt{2}}{2} \cdot 2 \log (2M)\right)^2 = \frac{32}{3-2\sqrt{2}} \log^2 (2M)
    \end{align*}

    Substituting this bound into our previous expression completes the proof.
    \begin{align*}
        \E\left[ L^2 \max_{m=1}^M \operatorname{diam}(Z_x^{m, \lambda} \cap \supp(\nu))^{2\beta} \right] & \leq L^2 \left(\E\left[ \max_{m=1}^M \operatorname{diam}(Z_x^{m, \lambda})^{2} \right]\right)^{\beta} \\
        & \leq L^2 \left( \frac{d}{\lambda^2} \cdot \E\left[ \left(\max_{m=1}^M W_{m}\right)^2 \right] \right)^{\beta} \\
        & \leq \left( \frac{32}{3-2\sqrt{2}} \right)^{\beta} \cdot \frac{L^2 d^{\beta}}{\lambda^{2\beta}} \cdot \log^2 (2M)
    \end{align*}
\end{proof}

\section{Derivation of a Moment Bound from a Tail Bound} \label{sc:moment_bound_from_tail_bound}

We derive a moment bound on the maximum of sub-exponential variables from a tail bound.

\begin{proof}[Derivation of \Cref{ln:max_zero_cell_diameter_tail_bound,ln:max_zero_cell_moment_bound_general} from \Cref{ln:zero_cell_diameter_tail_bound}]
    First, we justify \cref{ln:max_zero_cell_diameter_tail_bound} via the following numerical argument. If $dM$ is not too large, then the exponential term can absorb the coefficient with a small penalty.
    \[dM \leq \exp\left(\frac{c_\Pi t}{4}\right) \implies \sum_{m=1}^M \Prob\left[ \operatorname{diam}(Z_0^{m, 1}) \geq t \right] \leq 2dM e^{-c_\Pi \cdot \frac{t}{2}} \leq 2 \exp\left(-\frac{c_\Pi t}{6 \log (dM)}\right)\]
    The last inequality above holds because $dM \geq 2$. On the other hand, if $dM$ is very large, the claimed tail bound becomes trivial because it exceeds $1$.
    \[dM > \exp\left(\frac{c_\Pi t}{4}\right) \implies 2 \exp\left(-\frac{c_\Pi t}{6 \log (dM)}\right) > 2 \exp\left(-\frac{2}{3}\right) > 1\]

    Thus, \cref{ln:max_zero_cell_diameter_tail_bound} establishes a tail bound on $\max_{m=1}^M \operatorname{diam}(Z_0^{m, 1})$ for $dM \geq 2$, from which we now derive a bound on its $k$th moment for any $k \geq 1$, using the proof technique of Proposition 2.6.1 in \cite{Vershynin_2026}.
    \begin{align*}
        \E\left[ \left( \max_{m=1}^M \operatorname{diam}\left(Z_0^{m, 1}\right)\right)^k \right] & = \int_0^\infty \Prob\left[ \max_{m=1}^M \operatorname{diam}(Z_0^{m, 1}) \geq t \right] \cdot k t^{k-1} \; dt \\
        & \leq \int_0^\infty 2 \exp\left(-\frac{c_\Pi t}{6 \log (dM)}\right) \cdot k t^{k-1} \; dt \\
        & = 2k \left(\frac{6 \log (dM)}{c_\Pi}\right)^k \cdot \Gamma(k)
    \end{align*}
    The notation $\Gamma(\cdot)$ refers to the gamma function.

    For our purposes, it suffices to bound the second moment, i.e., $k = 2$.
    \begin{equation}
        \E\left[ \left( \max_{m=1}^M \operatorname{diam}\left(Z_0^{m, 1}\right)\right)^2 \right] \leq 144 \left(\frac{1}{c_\Pi}\right)^2 \log^2 (dM) \label{ln:max_zero_cell_second_moment_bound}
    \end{equation}
    
    Since $t \mapsto t^{\beta}$ is concave, we can apply Jensen's inequality to isolate the second moment and apply the bound from \cref{ln:max_zero_cell_second_moment_bound}.
    \begin{equation}
        \E\left[ \max_{m=1}^M \operatorname{diam}(Z_0^{m, 1})^{2\beta} \right] \leq \left(\E\left[ \left( \max_{m=1}^M \operatorname{diam}\left(Z_0^{m, 1}\right)\right)^{2} \right]\right)^{\beta} \leq  \frac{144^{\beta}}{c_\Pi^{2\beta}} \log^{2\beta} (dM) \label{ln:max_zero_cell_moment_bound}
    \end{equation}
    
    In the case of $d = M = 1$, a single tree's diameter $\operatorname{diam}(Z_0^{1, 1})$ is the maximum diameter $\max_{m=1}^M \operatorname{diam}(Z_0^{m, 1})$, so \cref{ln:zero_cell_diameter_tail_bound} immediately provides the tail bound we need to repeat the derivation of a moment bound.
    \begin{equation}
        \E\left[ \operatorname{diam}(Z_0^{1, 1})^{2\beta} \right] \leq \left(\E\left[ \left( \operatorname{diam}\left(Z_0^{1, 1}\right)\right)^{2} \right]\right)^{\beta} \leq 16^{\beta} \left(\frac{1}{c_\Pi}\right)^{2\beta} \label{ln:zero_cell_diameter_moment_bound}
    \end{equation}
\end{proof}

\section{Proof of Infinite $\mathcal{P}^M_\lambda$-transfer Discrepancy} \label{sc:proofs_example_shift}

\begin{proof}[Proof of \cref{pr:insufficient_ensemble_size}]
    We begin by expressing the integrated $\mathcal{P}^M_\lambda$-transfer discrepancy using the STIT tessellation's stationarity and eponymous property: stability under iteration \cite{oreilly_stochastic_2022}.
    \begin{equation}
    	\E\left[\frac{1}{\nu\left(\bigcup_{m=1}^M Z_X^{m, \lambda}\right)}\right] = \E\left[\frac{1}{\nu\left(\frac{1}{\lambda} \left( \bigcup_{m=1}^M Z_0^{m, 1} \right) + X\right)}\right] \label{ln:stit_zero_cell_union}
    \end{equation}

    Now, the independently realized random partitions are all Mondrian trees with lifetime $1$. The zero cell of the partition indexed $m$ is denoted $Z_0^{m, 1}$ and is distributed as the stochastic interval $[-T^-_m, T^+_m]$, where $T^-_m$ and $T^+_m$ are independent unit exponential random variables \cite{Mourtada_Gaiffas_Scornet_2020}. The union of these intervals is $[- \max_{m=1}^M T^-_m, \max_{m=1}^M  T^+_m]$. Hence, we can rewrite the integrated $\mathcal{P}^M_\lambda$-transfer exponent in terms of independent unit exponential random variables.
    \begin{equation}
    	\E\left[\frac{1}{\nu\left(\frac{1}{\lambda} \left( \bigcup_{m=1}^M Z_0^{m, 1} \right) + X\right)}\right] = \E\left[\frac{1}{\nu\left(\left[-\frac{\max_{m=1}^M T^-_m}{\lambda} + X, \frac{\max_{m=1}^M T^+_m}{\lambda} + X\right]\right)}\right] \label{ln:mondrian_zero_cell_union}
    \end{equation}
    
    Condition on the exponential random variables, and consider the expectation with respect to $X$ restricted to the case $X \leq \left(\max_{m=1}^M \frac{T^+_m}{\lambda}\right) \wedge 1$. This case is the most probable one because each $T^+_m$ has mean $1$. We restrict to this case only for convenience in the analysis.
    \[\E_X\left[\frac{1}{\nu\left(\left[-\frac{\max_{m=1}^M T^-_m}{\lambda} + X, \frac{\max_{m=1}^M T^+_m}{\lambda} + X\right]\right)}\right] \geq \E_X\left[\frac{\mathbf{1}_{X \leq \left(\max_{m=1}^M \frac{T^+_m}{\lambda}\right) \wedge 1}}{\nu\left(\left[-\frac{\max_{m=1}^M T^-_m}{\lambda} + X, \frac{\max_{m=1}^M T^+_m}{\lambda} + X\right]\right)}\right]\]

    We demonstrate that the right-hand side of the line above is at least on the order of $\frac{\lambda^\alpha}{(\max_{m=1}^M T^+_m)^\alpha}$. 
    Since $\nu$ has support $[0, 1]$, and in the event that $X \leq \left(\max_{m=1}^M \frac{T^+_m}{\lambda}\right) \wedge 1$,
    \[\nu\left(\left[-\frac{\max_{m=1}^M T^-_m}{\lambda} + X, \frac{\max_{m=1}^M T^+_m}{\lambda} + X\right]\right) \leq \nu\left(\left[0, \frac{\max_{m=1}^M T^+_m}{\lambda} + X\right]\right) \leq \nu\left(\left[0, \frac{\lambda + 1}{\lambda} \max_{m=1}^M T^+_m\right]\right)\]
    and substituting the density for $\nu$,
    \[\nu\left(\left[0, \frac{\lambda + 1}{\lambda} \max_{m=1}^M T^+_m\right]\right) = (\alpha + 1) \int_0^{\left( \frac{\lambda + 1}{\lambda} \max_{m=1}^M T^+_m \right) \wedge 1} x^\alpha \; dx = \left( \left( \frac{\lambda + 1}{\lambda} \max_{m=1}^M T^+_m \right) \wedge 1 \right)^{\alpha + 1}\]
    so that for all $\lambda \geq 1$,
    \begin{align*}
        \E_X\left[\frac{\mathbf{1}_{X \leq \left( \max_{m=1}^M T^+_m \right) \wedge 1}}{\nu\left(\left[-\frac{\max_{m=1}^M T^-_m}{\lambda} + X, \frac{\max_{m=1}^M T^+_m}{\lambda} + X\right]\right)}\right] & = \int_0^{\left( \max_{m=1}^M T^+_m \right) \wedge 1} \frac{dx}{\nu\left(\left[-\frac{\max_{m=1}^M T^-_m}{\lambda} + x, \frac{\max_{m=1}^M T^+_m}{\lambda} + x\right]\right)} \\
        & \geq \int_0^{\left( \max_{m=1}^M T^+_m \right) \wedge 1} \frac{dx}{\left( \left( \frac{\lambda + 1}{\lambda} \max_{m=1}^M T^+_m \right) \wedge 1 \right)^{\alpha + 1}} \\
        & = \frac{\left( \max_{m=1}^M T^+_m \right) \wedge 1}{\left( \left( \frac{\lambda + 1}{\lambda} \max_{m=1}^M T^+_m \right) \wedge 1 \right)^{\alpha + 1}} \\
        & \geq \frac{1}{2^{\alpha + 1}} \cdot \frac{1}{\left( \left( \max_{m=1}^M T^+_m \right) \wedge 1 \right)^\alpha}
    \end{align*}

    Note that the maximum of independent exponential random variables $\{T^+_m\}_{m=1}^M$ is less than their sum, and their sum is a Gamma-distributed random variable with shape parameter $M$ and rate parameter $1$.
    \[\left( \left( \max_{m=1}^M T^+_m \right) \wedge 1 \right) \leq \max_{m=1}^M T^+_m \leq \sum_{m=1}^M T^+_m \sim \operatorname{Gamma}(M, 1)\]

    The first inverse moment of the unit exponential random variable is infinite by comparison to the harmonic series on the interval $(0, 1)$. From this, we extrapolate that the $\operatorname{Gamma}(M, 1)$ variable's moment of order $-\alpha$ is infinite for $\alpha \geq M$.
    \begin{align*}
        \E_{\{T^+_m\}_m} \left[\frac{1}{\left(\sum_{m=1}^M T^+_m\right)^\alpha}\right] & = \int_0^\infty \frac{1}{t^\alpha} \cdot \frac{1}{\Gamma(M)} t^{M-1} e^{-t} \; dt \\
        & = \frac{1}{\Gamma(M)} \int_0^\infty \frac{1}{t^{\alpha - M + 1}} \cdot e^{-t} \; dt \\
        & \geq \frac{1}{\Gamma(M)} \int_0^1 \frac{1}{t^{\alpha - M + 1}} \cdot e^{-t} \; dt \\
        & \geq \frac{1}{\Gamma(M)} \int_0^1 \frac{1}{t} \cdot e^{-t} \; dt \\
        & = \infty
    \end{align*}

    Hence, the integrated $\mathcal{P}^M_\lambda$-transfer exponent is infinite.
    \begin{align*}
        \E\left[\frac{1}{\nu\left(\bigcup_{m=1}^M Z_X^{m, \lambda}\right)}\right] & \geq \E_{\{T^-_m, T^+_m\}_m} \E_X\left[\frac{\mathbf{1}_{X \leq \left(\max_{m=1}^M \frac{T^+_m}{\lambda}\right) \wedge 1}}{\nu\left(\left[-\frac{\max_{m=1}^M T^-_m}{\lambda} + X, \frac{\max_{m=1}^M T^+_m}{\lambda} + X\right]\right)}\right] \\
        & \geq \E_{\{T^+_m\}_m} \left[ \frac{1}{2^{\alpha + 1}} \cdot \frac{1}{\left( \left( \max_{m=1}^M T^+_m \right) \wedge 1 \right)^\alpha} \right] \\
        & \geq \frac{1}{2^{\alpha + 1}} \cdot \E_{\{T^+_m\}_m} \left[\frac{1}{\left(\sum_{m=1}^M T^+_m\right)^\alpha}\right] \\
        & = \infty
    \end{align*}
\end{proof}

\bibliographystyle{abbrv}
\bibliography{gbo_syllabus,references,other}

\end{document}